\documentclass{article}

\usepackage{arxiv}

\usepackage[utf8]{inputenc} 
\usepackage[T1]{fontenc}    
\usepackage{hyperref}       
\usepackage{url}            
\usepackage{booktabs}       
\usepackage{amsfonts}       
\usepackage{nicefrac}       
\usepackage{microtype}      
\usepackage{lipsum}		
\usepackage{graphicx}
\usepackage{natbib}
\usepackage{doi}
\usepackage{amsmath,amssymb}

\newcommand{\mat}[1]{\boldsymbol{#1}}
\newcommand{\bref}[1]{(\ref{#1})}
\newcommand{\set}[1]{\mathcal{#1}}

\newcommand{\rmvec}{\mathsf{vec}}
\newcommand{\diag}{\mathsf{diag}}
\newcommand{\ul}[1]{\underline{#1}}

\newcommand{\barmat}[1]{\bar{\boldsymbol{#1}}}

\title{Gaussian Process Regression With~Local~Explanation}

\author{Yuya Yoshikawa\thanks{This work has been submitted to the IEEE for possible publication. Copyright may be transferred without notice, after which this version may no longer be accessible.}
\\
	Software Technology and Artificial Intelligence Research Laboratory\\
	Chiba Institute of Technology\\
	\texttt{yoshikawa@stair.center} \\
	\And
	Tomoharu Iwata \\
	NTT Communication Science Laboratories\\
	\texttt{tomoharu.iwata.gy@hco.ntt.co.jp} \\
}

\renewcommand{\shorttitle}{Gaussian Process Regression With~Local~Explanation}

\hypersetup{
pdftitle={Gaussian Process Regression With~Local~Explanation},
pdfauthor={Yuya Yoshikawa, Tomoharu Iwata},
pdfkeywords={Gaussian processes, interpretable machine learning, locally linear models, explainability, feature relevance},
}

\begin{document}
\maketitle

\begin{abstract}
Gaussian process regression (GPR) is a fundamental model used in machine learning.
Owing to its accurate prediction with uncertainty and versatility in handling various data structures via kernels, GPR has been successfully used in various applications.
However, in GPR, how the features of an input contribute to its prediction cannot be interpreted.
Herein, we propose GPR with local explanation, which reveals the feature contributions to the prediction of each sample, while maintaining the predictive performance of GPR.
In the proposed model, both the prediction and explanation for each sample are performed using an easy-to-interpret locally linear model.
The weight vector of the locally linear model is assumed to be generated from multivariate Gaussian process priors.
The hyperparameters of the proposed models are estimated by maximizing the marginal likelihood.
For a new test sample, the proposed model can predict the values of its target variable and weight vector, as well as their uncertainties, in a closed form.
Experimental results on various benchmark datasets verify that the proposed model can achieve predictive performance comparable to those of GPR and superior to that of existing interpretable models, and can achieve higher interpretability than them, both quantitatively and qualitatively.    
\end{abstract}

\section{Introduction}\label{sec:intro}
Gaussian processes (GPs) have been well studied for constructing probabilistic models as priors of nonlinear functions in the machine learning (ML) community. They have demonstrated great success in various problem settings, such as regression~\citep{rasmussen2003gaussian,wilson2012gaussian}, classification~\citep{rasmussen2003gaussian,csato2000efficient}, time-series forecasting~\citep{roberts2013gaussian}, and black-box optimization~\citep{snoek2012practical}.
A fundamental model on GPs is Gaussian process regression (GPR)~\citep{rasmussen2003gaussian}; owing to its high predictive performances and versatility in using various data structures via kernels, it has been used in not only the ML community, but also in various other research areas, such as finance~\citep{gonzalvez2019financial}, geostatistics~\citep{camps2016survey}, material science~\citep{zhang2020bayesian} and medical science~\citep{cheng2017sparse,futoma2018gaussian}.

GPR is defined on an infinite-dimensional feature space via kernel functions.
Therefore, it requires the values of the kernels defined on pairs of samples, i.e., a covariance matrix of the samples as an input, rather than the samples themselves.
Owing to the nonlinearity of the kernel, GPR enables nonlinear predictions.
In terms of interpretability, the covariance is useful for understanding the relationship between the samples; however, since the kernels make the features invisible, GPR cannot explain which features contribute to the predictions, like linear regression models.
Therefore, it prevents us from judging whether the predictions by GPR are reasonable and performed by fair decision.

For the interpretability of ML, several methodologies that explain the features that contribute to the outputs of prediction models, including GPR, have been proposed; in this case, the prediction models are often regarded as black-boxes~\citep{molnar2019,chen2018learning}.
Their representative methods are local interpretable model-agnostic explanations (LIME)~\citep{ribeiro2016} and Shapley additive explanations (SHAP)~\citep{lundberg2017}, which approximate the prediction for each test sample by a locally linear explanation model.
Since the weights of the learned explanation model represent feature contributions to the prediction, they can assist ML practitioners and scientists to understand the behavior of the prediction models and the functionality of the features in the prediction.
However, some limitations exist in these methods.
First, because the forms of the prediction and explanation models differ, it is unsure whether the estimated feature contributions reflect those of the prediction model.
Second, because the explanation model is learned on each test sample, it may not obtain consistent explanations on similar samples.

\begin{figure}[t]
\centering
\includegraphics[keepaspectratio,width=70mm]{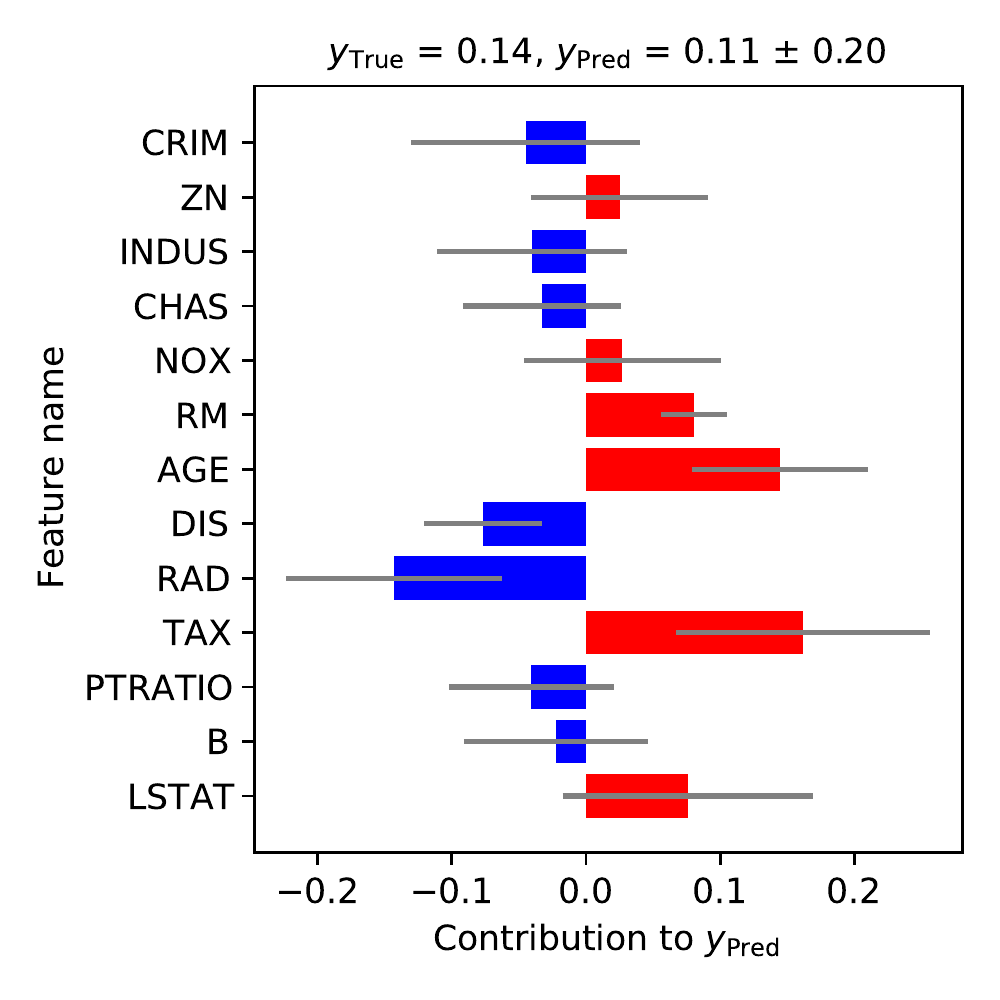}
\caption{
Example of explanation for prediction by GPX for a sample on the Boston housing dataset~\citep{harrison1978hedonic}.
Red and blue bars indicate positive and negative predicted feature contributions, respectively, and error bars indicates predicted standard deviation of the feature contributions.
We provide further examples and feature description in Appendix~\ref{sec:appendix:boston}.}
\label{fig:example}
\end{figure}

To overcome the aforementioned limitations, we propose a novel framework for GP-based regression models, {\it \ul{G}aussian \ul{p}rocess regression with local e\ul{x}planation}, called {\it GPX}, which reveals the feature contributions to the prediction for each sample, while maintaining the predictive performance of GPR.
In GPX, both the prediction and explanation for each sample are performed using an easy-to-interpret locally linear model.
Therefore, no gap exists between the prediction and explanation.
The weight vector of the locally linear model is assumed to be generated from multivariate GP priors~\citep{alvarez2012kernels}.
As the multivariate GP priors have a covariance function defined as kernels on the samples, GPX ensures that similar samples have similar weights.
The hyperparameters of GPX are estimated by maximizing the marginal likelihood, in which the weight vectors for all the training samples are integrated out.
For a test sample, the predictions with their uncertainties of the target variable and weight vector are obtained by computing their predictive distributions.
The explanation for the predicted target variable is provided using the estimated weight vector with uncertainty, as shown in Figure~\ref{fig:example}.
Depicting the explanation with uncertainty helps users of GPX judge the reasonability of the predicted weights. 

In experiments, we evaluated GPX both qualitatively and quantitatively in terms of predictive performance and interpretability on various benchmark datasets.
The experimental results show that 1) GPX can achieve predictive errors comparable to GPR and lower errors compared with existing interpretable methods, 2) it can outperform model-agnostic interpretable methods and locally linear methods in terms of three interpretability measurements, and 3) the feature contributions produced by GPX are appropriate.

\section{Related Work}\label{sec:RW}
Linear regression models are simple types of interpretable models, as their weight for each feature directly represents the contribution to the output of the models if the feature has a binary value, where the contribution helps us to interpret the effectiveness of the feature on a task.
For real-valued features, the contribution for each feature is calculated as the product of the feature value and its corresponding weight.
A number of studies introducing various regularizations have been conducted to produce methods such as the ridge regression~\citep{hoerl1970} and lasso~\citep{tibshirani1996} methods.
These models have a {\it global} weight vector that is shared across all samples.
In kernel methods, automatic relevance determination (ARD), which considers the global relevance of each feature contained in kernel functions, is widely used~\citep{neal2012bayesian,wipf2008new}.
The above approaches are beneficial in understanding global effectiveness of features.
However, since linear regression models often does not fit real complicated problems, resulting in low predictive accuracy, the estimated weights may be also unreliable.
In addition, these approaches cannot estimate weights/relevances appropriate for individual samples, which means that they cannot cope with the need of the significant changes of the weights/relevances among the samples.
For example, as shown in the result on Digits dataset (Figure~\ref{fig:experiment:digits}), this is crucial in image classification tasks that important pixels (i.e., features) and their weights change depending on individual images (i.e., samples).

On the other hand, some {\it locally} linear models for regression have been proposed, such as the network lasso~\citep{hallac2015} and localized lasso~\citep{yamada2017}, which have a weight vector for each sample.
Therefore, these methods can avoid the drawbacks of globally linear models.
To receive the benefit, we focus on generating predictions with explanations using locally linear models.
In the network and localized lasso, the weights of locally linear models are estimated via optimization with network-based regularization, where the network must be defined on samples in advance.
If the network is not provided, as assumed in standard regression problems, we can construct a $k$-nearest-neighbor graph of samples to create a network, where $k$ is a hyperparameter that must be optimized via cross validation.
Meanwhile, GPX can estimate weights and their uncertainties without constructing graphs by assuming that weights are determined by functions generated from GPs.


For GP models, \citep{paananen2019variable} proposed feature selection methods that quantify the feature relevances by measuring the difference between the predictive posterior for each sample and that for its vicinity.
The feature relevance can be used for evaluating the importance of the feature in prediction.
Meanwhile, the feature contribution produced by GPX indicates whether the feature makes a positive or negative contribution to an individual prediction and how much it contributes.

With regard to research on deep neural networks (DNNs), a number of studies have been conducted on making predictions generated by DNNs interpretable~\citep{chen2019looks,arras2017relevant,ying2019gnnexplainer}.
Some of these studies have developed methods that make interpretable predictions by generating locally linear models for each sample using DNNs~\citep{melis2018,schwab2019,yoshikawa2020neural}.
These concepts inspired our study, but we formalize our model without DNNs.
To the best of our knowledge, our study is the first to develop a GP-based regression model with local explanations.
Compared to the DNN-based locally linear models, GPX has mainly two benefits: 1) GPX can produce the feature contributions with uncertainty, and 2) GPX can be straightforwardly used for tasks in which GPR can be used advantageously, such as Bayesian optimization~\citep{snoek2012practical,golovin2017google}.

\citep{wilson2012gaussian} proposed Gaussian process regression networks, which have similar structure with multi-layer perceptron.
Herein, the weights of the networks are generated from GPs.
Although their idea is related to GPX, their work did not focus on improving the explainability of the predictions by GPR.

\section{Proposed Model}\label{sec:proposed}
In this section, we describe the proposed model, i.e., {\it Gaussian process regression with local explanation}, called {\it GPX}.

We consider a scalar-valued regression problem.
Suppose that training data $\set{D} = \{ (\mat{x}_i, y_i, \mat{z}_i) \}_{i=1}^n$ containing $n$ samples is provided.
$\mat{x}_i \in \set{X}$ is an original input representing the $i$th sample, where $\set{X}$ is an original input space.
Although a typical representation for $\mat{x}_i$ is a vector, it can be any data representation on which kernel functions are defined, such as graphs~\citep{vishwanathan2010graph} and sets~\citep{muandet2012learning,yoshikawa2014latent}.
$y_i \in \mathbb{R}$ is a target variable for the sample.
$\mat{z}_i \in \mathbb{R}^d$ is a $d$-dimensional vector of simplified representation for $\mat{x}_i$.
Because GPX explains the prediction via a simplified representation, the meaning of each dimension of $\mat{z}_i$ should be easily understood by humans, e.g., tabular data and bag-of-words representation for text.
$\mat{z}_i$ is an optional input; therefore, if $\mat{x}_i$ can be used as a simplified representation, one can define $\mat{z}_i = \mat{x}_i$.
Let us denote $\mat{X} = \{\mat{x}_i\}_{i=1}^n$, $\mat{y} = (y_i)_{i=1}^n \in \mathbb{R}^n$ and $\mat{Z} = (\mat{z}_i)_{i=1}^n \in \mathbb{R}^{n \times d}$.

In GPX, both the prediction of target variables $\mat{y}$ and their explanations are performed via easy-to-interpret locally linear models, i.e., target variable $y_i$ for the $i$th sample is assumed to be obtained using locally linear function $f_i: \mathbb{R}^d \rightarrow \mathbb{R}$, defined as follows:
\begin{equation}
f_i(\mat{z}_i) = \mat{w}_i^\top \mat{z}_i + \epsilon_\mathrm{y},
\end{equation}
where $\mat{w}_i \in \mathbb{R}^d$ is a $d$-dimensional weight vector for the $i$th sample, and $\epsilon_\mathrm{y} \sim \mathcal{N}(0, \sigma_\mathrm{y}^2)$ is a Gaussian noise with variance $\sigma_\mathrm{y}^2 > 0$.
Here, the explanation for the $i$th sample is obtained using either weight vector $\mat{w}_i$ or feature contributions $\mat{\phi}_i = (w_{il} z_{il})_{l=1}^d$.

Estimating $\mat{W} = (\mat{w}_i)_{i=1}^n \in \mathbb{R}^{n \times d}$ without any constraints is an ill-posed problem because the number of free parameters in $\mat{W}$, $nd$, is larger than that of target variable $n$.
To avoid this problem in GPX, we assume that functions determining $\mat{W}$ are generated from a multivariate GP.
More specifically, weight vector $\mat{w}_i$ for the $i$th sample is obtained as follows:
\begin{equation}
\mat{w}_{i} = \mat{g}(\mat{x}_i) + \mat{\epsilon}_\mathrm{w},
\end{equation}
where $\mat{\epsilon}_\mathrm{w} \sim \mathcal{N}(\mat{0}, \sigma_\mathrm{w}^2 \mat{I}_d)$ is a $d$-dimensional Gaussian noise with variance $\sigma_\mathrm{w}^2 > 0$, and $\mat{I}_d$ is an identity matrix of order $d$.
Here, vector-valued function $\mat{g}: \set{X} \rightarrow \mathbb{R}^d$ is a function that determines the weight vector for each sample, and each element of $\mat{g}$ is generated from a univariate GP independently, as follows:
%
\begin{align}
\mat{g}(\mat{x}) &= \left(\ g_1(\mat{x}),g_2(\mat{x}),\cdots,g_d(\mat{x})\ \right)^\top, \\
\intertext{where}
g_l(\mat{x}) \ &\sim \ \mathcal{GP}(m(\mat{x}), k_\theta(\mat{x}, \mat{x}')),
\end{align}
where $m(\mat{x})$ is the mean function, and $k_{\mat{\theta}}(\mat{x},\mat{x}')$ is the covariance function with set of parameters $\mat{\theta}$. 
Herein, we use zero mean function for $m(\mat{x})$.
Covariance function $k_\theta(\mat{x},\mat{x}')$ is a kernel function defined on two inputs $\mat{x}$, $\mat{x}' \in \set{X}$. 
For example, one can use a scaled RBF kernel with parameters $\mat{\theta} = \{\theta_1,\theta_2\}$ as the kernel function when $\mat{x}, \mat{x}'$ are vectors, defined as follows:
\begin{equation}
k_{\mat{\theta}}(\mat{x},\mat{x}') = \theta_1 \exp\left( -\frac{1}{\theta_2} \|\mat{x}-\mat{x}'\|^2_2 \right)
\quad (\theta_1, \theta_2 > 0).
\label{eq:rbf}
\end{equation}
By using $\mat{g}$ generated as such, GPX ensures that two similar samples, i.e., those having a large kernel value, have similar weight vectors.

We let $\mat{G} = (\mat{g}(\mat{x}_i)^\top)_{i=1}^{n} \in \mathbb{R}^{n \times d}$.
Based on the generative process above, the joint distribution of GPX is written as follows:
%
\begin{align}
\label{eq:joint_distribution}
p(\mat{y}, \mat{W}, \mat{G} \mid \mat{X}, \mat{Z})
= p(\mat{G} \mid \mat{X}) \prod_{i=1}^n p(y_i \mid \mat{w}_i,\mat{z}_i) p(\mat{w}_i \mid \mat{G}_{i,\cdot}),
\end{align}
where
\begin{align}
p(y_i \mid \mat{w}_i,\mat{z}_i) &= \mathcal{N}(y_i \mid \mat{w}_i^\top \mat{z}_i, \sigma_\mathrm{y}^2),\\
p(\mat{w}_i \mid \mat{G}_{i,\cdot}) &= \mathcal{N}(\mat{w}_i \mid \mat{g}(\mat{x}_i), \sigma_\mathrm{w}^2 \mat{I}_d),\\
p(\mat{G} \mid \mat{X}) &= \prod_{l=1}^d \mathcal{N}\left(\mat{G}_{\cdot,l} \mid \mat{0}, \mat{K} \right), 
\end{align}
where $\mat{G}_{i,\cdot}$ and $\mat{G}_{\cdot,l}$ denote the $i$th row and $l$th column vectors of $\mat{G}$, respectively, and $\mat{K} = \left(k_{\mat{\theta}}(\mat{x}_i, \mat{x}_j) \right)_{i,j=1}^n$ is a Gram matrix of order $n$, which is identical to the requirement in GPR.

\section{Training and Prediction}
\label{sec:train}
In this section, we describe the derivation of the marginal likelihood of GPX, the hyperparameter estimation for GPX, and the derivation of the predictive distributions of target variables and weight vectors for test samples.

\subsubsection*{Marginal likelihood}
To ease the derivation of the marginal likelihood, we first modified the formulation of the joint distribution~\bref{eq:joint_distribution}, while maintaining its mathematical meanings, as follows:
%
\begin{align}
\label{eq:joint_distribution2}
\lefteqn{p(\mat{y}, \mat{W}, \mat{G} \mid \mat{X}, \mat{Z})}\quad \\
&=\ \mathcal{N}(\rmvec(\mat{G}) \mid \mat{0}, \barmat{K})
\ \mathcal{N}(\rmvec(\mat{W}) \mid \rmvec(\mat{G}), \sigma_\mathrm{w}^2 \mat{I}_{nd}) 
\times \mathcal{N}(\mat{y} \mid \barmat{Z} \rmvec(\mat{W}), \sigma_\mathrm{y}^2 \mat{I}_{nd}), \nonumber
\end{align}
where $\bar{\mat{K}}$ is a block diagonal matrix of order $nd$ whose block is $\mat{K}$, and $\rmvec(\cdot)$ is a function that flattens the input matrix in a column-major order.
Here, 
\begin{equation}
\barmat{Z} = (\diag(\mat{Z}_{\cdot,1}), \diag(\mat{Z}_{\cdot,2}), \cdots, \diag(\mat{Z}_{\cdot,d})) \in \mathbb{R}^{n \times nd},
\end{equation}
%
%
where $\diag(\cdot)$ is a diagonal matrix whose diagonal elements possess the values of the input vector.
In~\bref{eq:joint_distribution}, $d$ functions that output $n$-dimensional column vectors in $\mat{W}$ are generated from GPs; however, in~\bref{eq:joint_distribution2}, it is rewritten such that a single function that outputs an $nd$-dimensional flatten vector $\rmvec(\mat{W})$ is generated from a single GP.
Consequently, the likelihood of target variables $\mat{y}$ can be rewritten as a single multivariate normal distribution.

Subsequently, we derived the marginal likelihood by integrating out $\mat{G}$ and $\mat{W}$ in~\bref{eq:joint_distribution2}.
Owing to the property of normal distributions, it can be obtained analytically, as follows:
%
\begin{align}
p(\mat{y} \mid \mat{X}, \mat{Z}) 
&= \iint p(\mat{y}, \mat{W}, \mat{G} \mid \mat{X}, \mat{Z}) d\mat{W} d\mat{G} \\
&= \mathcal{N}\left(\mat{y} \mid \mat{0},\mat{C} \right), \nonumber
\label{eq:marginal}
\end{align}
where 
\begin{align}
\mat{C} 
&= \sigma_\mathrm{y}^2\mat{I}_n + \bar{\mat{Z}}\left(\bar{\mat{K}} + \sigma_\mathrm{w}^2 \mat{I}_{nd}\right) \bar{\mat{Z}}^\top \\ 
&= \sigma_\mathrm{y}^2\mat{I}_n + (\mat{K}  + \sigma_\mathrm{w}^2 \mat{I}_n) \odot \mat{Z}\mat{Z}^\top. \nonumber
\end{align}


\subsubsection*{Hyperparameter estimation}
If $k_\theta(\mat{x},\mat{x}')$ is differentiable with respect to $\theta$, all the hyperparameters, i.e., $\theta$, $\sigma_\mathrm{w}$, and $\sigma_\mathrm{y}$, can be estimated by maximizing the logarithm of the marginal likelihood with respect to them for the training data using gradient-based optimization methods, e.g., L-BFGS~\citep{liu1989limited}.

\subsubsection*{Predictive distributions}
For a new test sample $(\mat{x}_*, \mat{z}_*)$, our goal is to infer the predictive distributions of target variable $y_*$ and weight vector $\mat{w}_*$.
First, the predictive distribution of $y_*$ is obtained similarly as in the standard GPR, as follows:
\begin{equation}
\label{eq:predictive_y}
p(y_* \mid \mat{x}_*,\mat{z}_*,\set{D})
= \mathcal{N}\left(y_* \mid \mat{c}_*^\top\mat{C}^{-1}\mat{y}, c_{**} - \mat{c}_*^\top \mat{C}^{-1} \mat{c}_* \right),
\end{equation}
where $\mat{c}_* = \left(k_\theta(\mat{x}_*, \mat{x}_i) \mat{z}_*^\top \mat{z}_i \right)_{i=1}^n \in \mathbb{R}^n$ and $c_{**} = \sigma_\mathrm{y}^2 + \left(k_\theta(\mat{x}_*, \mat{x}_*) + \sigma_\mathrm{w}^2 \right)\mat{z}_*^\top \mat{z}_*  \in \mathbb{R}$.

Second, the predictive distribution of $\mat{w}_*$ is obtained by solving the following integral:
%
%
%
\begin{align}
p(\mat{w}_* \mid \mat{x}_*,\mat{z}_*,\set{D}) 
&= \int p(\mat{w}_* \mid \mat{W},\mat{X},\mat{x}_*) p(\mat{W} \mid \set{D}) d\mat{W}, \\
p(\mat{w}_* \mid \mat{W},\mat{X},\mat{x}_*)
&= \mathcal{N}(\mat{w}_* \mid \mat{A}\rmvec(\mat{W}), \barmat{c}_{**} - \mat{A}\barmat{k}_* ), \\
p(\mat{W} \mid \set{D})
&= \mathcal{N}(\rmvec(\mat{W}) \mid \sigma_\mathrm{y}^{-2} \mat{S}\barmat{Z}^\top \mat{y}, \mat{S}),
\end{align}
where we define $\mat{A} = \barmat{k}_*^\top\left(\barmat{K} + \sigma_\mathrm{w}^2 \mat{I}_{nd} \right)^{-1}$, $\mat{S} = \mat{L} - \mat{L}\barmat{Z}^\top \mat{C}^{-1} \barmat{Z}\mat{L}$, $\mat{L} = \barmat{K} + \sigma_\mathrm{w}^2 \mat{I}_{nd}$, and $\barmat{c}_{**} = \left(k_\theta(\mat{x}_*,\mat{x}_*) + \sigma_\mathrm{w}^2 \right)\mat{I}_d$.
$\barmat{k}_*$ is an $nd$-by-$d$ block matrix, where each block is an $n$-by-$1$ matrix, and $(l,l)$-block of the block matrix is $(k_\theta(\mat{x}_*,\mat{x}_i))_{i=1}^n$ for $l=1,2,\cdots,d$, and the other blocks are zero matrices.
Solving the integral analytically according to the property of the normal distributions, we obtain 
%
\begin{align}
\label{eq:predictive_w}
p(\mat{w}_* \mid \mat{x}_*,\mat{z}_*,\set{D})
= \mathcal{N}(\mat{w}_* \mid \sigma_\mathrm{y}^{-2} \mat{A} \mat{S} \barmat{Z}^\top \mat{y}, \barmat{c}_{**} - \mat{A}\barmat{k}_* + \mat{A}\mat{S}\mat{A}^\top). 
\end{align}
We provide the detailed derivation of predictive distributions \bref{eq:predictive_y} and \bref{eq:predictive_w} in Appendix~\ref{sec:appendix:derivation}.

The marginal likelihood~\bref{eq:marginal} and the predictive distribution for $y_*$~\bref{eq:predictive_y} are similar to those of GPR, except that GPX can obtain the predictive distribution for $\mat{w}_*$~\bref{eq:predictive_w}.
Since GPX can be used with the same input as GPR if $\mat{Z} = \mat{X}$, it can be employed in existing ML models, instead of GPR.

\subsubsection*{Computational efficiency}
As with ordinary GPR, the computational cost of GPX is dominated by the inverse computation.
The computation of $\mat{A}$ requires inverting a square matrix of order $nd$, $\barmat{K} + \sigma_\mathrm{w}^2 \mat{I}_{nd}$.
However, because the matrix is block diagonal and every diagonal block comprises $\mat{K} + \sigma_\mathrm{w}^2\mat{I}_n$, a square matrix of order $n$, $\mat{A}$ can be obtained by inverting $\mat{K} + \sigma_\mathrm{w}^2\mat{I}_n$ only once.
The remaining inverse matrix $\mat{C}^{-1}$ is of order $n$.
Therefore, all the inverse matrices appearing in GPX can be obtained using a naive implementation with a computational complexity of $\mathcal{O}(n^3)$, which is the same as that in GPR.
To significantly reduce the computational cost, efficient computation methods for GPR, such as the inducing variable method~\citep{titsias2009variational} and KISS-GP~\citep{wilson2015kernel}, can be used for GPX.
In addition, because $\barmat{k}_*, \barmat{K}$ and $\barmat{Z}$ are sparse matrices, one can obtain the predictive distributions efficiently using libraries for sparse matrix computation.

\section{Experiments}\label{sec:experiment}
In this section, we demonstrate the effectiveness of the proposed model, GPX, quantitatively and qualitatively, by comparing various interpretable models.
Through a quantitative evaluation, we evaluated the models based on the following perspectives:
%
\begin{itemize}
\item \textbf{Accuracy:} How accurate is the prediction of the interpretable model?
\item \textbf{Faithfulness:} Are feature contributions indicative of ``true'' importance?
\item \textbf{Sufficiency:} Do $k$-most important features reflect the prediction?
\item \textbf{Stability:} How consistent are the explanations for similar or neighboring examples?
\end{itemize}
%
In addition, we qualitatively evaluated whether the feature contributions produced by the models were appropriate by visualizing them.
Subsequently, we experimentally compared the computational efficiency of the models.

All the experiments were done with a computer with Intel Xeon Gold 6132 2.6GHz CPU with 16 cores, and 120GB of main memory.

\subsection{Preparation}
\subsubsection*{Datasets}
We used eight datasets in the UCI machine learning repository~\citep{dua2019}, referred to as Digits, Abalone, Diabetes, Boston, Fish, Wine, Paper and Drug in our experiments.
We provide the details of the datasets in Appendix~\ref{sec:appendix:dataset}.
Digits dataset is originally a classification dataset for recognizing handwritten digits from 0 to 9.
To use it as a regression problem, we transformed the labels into target variables $\mat{y}$ of scalar values, i.e., the target variables for the labels from 0 to 4 were $-1$, and those for the remaining labels were $1$.
With Paper and Drug datasets whose samples were represented as sentences, the original input $\mat{X}$ and the simplified input $\mat{Z}$ differed, i.e., we used the 512-dimensional sentence vectors obtained using Sentence Transformers~\citep{reimers2020making} as $\mat{X}$, while we used bag-of-words binary vectors for the sentences as $\mat{Z}$.
Each of the remaining datasets had the same $\mat{X}$ and $\mat{Z}$.
In all the datasets, the values of $\mat{X}$ and $\mat{y}$ were standardized before training and prediction.
For a quantitative evaluation of each dataset, we evaluated the average scores over five experiments performed on different training/test splittings, where the training set was 80\% of the entire dataset, whereas the remaining was the test set.

\subsubsection*{GPX setup}
In GPX, we consistently used a scaled RBF kernel defined as~\bref{eq:rbf}.
The hyperparameters of GPX were estimated based on the method described in Section~\ref{sec:train}, where they were initialized with $\theta_1 = 1.0$, $\sigma_\mathrm{y} = 0.1$ and $\sigma_\mathrm{w} = 0.1$.
In addition, we initialized bandwidth parameter $\theta_2$ using median heuristics~\citep{garreau2017large}.

\subsubsection*{Comparing methods}
We compared GPX with several methods with globally or locally linear weights that can used as interpretable feature contributions or relevances for predictions.
Lasso~\citep{tibshirani1996} and Ridge~\citep{hoerl1970} are standard linear regression models with $\ell_1$ and $\ell_2$ regularizers, respectively, where their weights are globally shared across all samples.
ARD is a GPR model with ARD kernel that identifies the relevance of each feature.
The network lasso (``Network'' for short) is a locally linear model that regularizes the weights of nodes such that neighboring nodes in a network have similar weights~\citep{hallac2015}.
In our case, each node represents a sample, and the network is a $k$-nearest neighbor graph on the samples based on the cosine similarity on $\mat{X}$.
The localized lasso (``Localized'' for short) is an extension of the network lasso; it can estimate the sparse and exclusive weights of each sample by further incorporating an $\ell_{1,2}$ regularizer into the network lasso~\citep{yamada2017}.

To compare model-agnostic interpretable methods with GPX in terms of explainablity for prediction, we used LIME~\citep{ribeiro2016} and Kernel SHAP~\citep{lundberg2017}, which produce a locally linear model for each test sample to explain the prediction by a black-box prediction model.
For a fair comparison, we used GPR with RBF kernel as the prediction model.
In addition, we used a Kullback-Leibler (KL) divergence-based feature selection method for GPR with ARD kernel (``KL'' for short)~\citep{paananen2019variable}.
The hyperparameters of GPR were estimated by maximizing marginal likelihood, similarly as for GPX.
Meanwhile, those of the remaining comparing methods were optimized by grid search.
We provide the detailed description of the comparing methods in Appendix~\ref{sec:appendix:method}.

\subsection{Results}

\begin{table*}[t]
\centering
\caption{
Average and standard deviation of mean squared errors (MSEs) for the predictions of target variables on each dataset (lower scores are better).
Values in bold typeface are not statistically different (at 5\% level) from the best performing method in each row according to a paired t-test.
As GPR is not an interpretable model, we only included its performances in this table to show that those of GPX and GPR are similar.
}
\label{tab:experiment:accuracy}
\resizebox{\textwidth}{!}{%
\begin{tabular}{@{}rrrrrrr|r@{}}
\toprule
& GPX (ours)              & Localized         & Network           & ARD               & Lasso             & Ridge             & GPR               \\ \midrule
Digits   & {\bf 0.078 $\pm$ 0.010} & 0.135 $\pm$ 0.042 & 0.163 $\pm$ 0.033 & 0.258 $\pm$ 0.371 & 0.399 $\pm$ 0.028 & 0.398 $\pm$ 0.024 & 0.074 $\pm$ 0.008 \\
Abalone  & {\bf 0.428 $\pm$ 0.036} & 0.519 $\pm$ 0.023 & 0.534 $\pm$ 0.016 & {\bf 0.428 $\pm$ 0.026} & 0.477 $\pm$ 0.052 & 0.477 $\pm$ 0.053 & 0.427 $\pm$ 0.034 \\
Diabetes & {\bf 0.493 $\pm$ 0.041} & 0.610 $\pm$ 0.062 & 0.667 $\pm$ 0.091 & {\bf 0.492 $\pm$ 0.042} & 0.504 $\pm$ 0.039 & 0.503 $\pm$ 0.040 & 0.490 $\pm$ 0.048 \\
Boston   & {\bf 0.116 $\pm$ 0.053} & 0.208 $\pm$ 0.070 & 0.233 $\pm$ 0.062 & {\bf 0.115 $\pm$ 0.047} & 0.293 $\pm$ 0.078 & 0.284 $\pm$ 0.081 & 0.116 $\pm$ 0.052 \\
Fish     & {\bf 0.370 $\pm$ 0.061} & 0.479 $\pm$ 0.051 & 0.523 $\pm$ 0.054 & {\bf 0.376 $\pm$ 0.065} & 0.437 $\pm$ 0.055 & 0.437 $\pm$ 0.055 & 0.375 $\pm$ 0.066 \\
Wine     & {\bf 0.579 $\pm$ 0.048} & 0.715 $\pm$ 0.024 & 0.908 $\pm$ 0.051 & {\bf 0.628 $\pm$ 0.053} & 0.723 $\pm$ 0.039 & 0.712 $\pm$ 0.044 & 0.605 $\pm$ 0.046 \\
Paper    & {\bf 0.806 $\pm$ 0.054} & 0.981 $\pm$ 0.058 & 0.919 $\pm$ 0.088 & {\bf 0.815 $\pm$ 0.092} & 0.821 $\pm$ 0.047 & 0.936 $\pm$ 0.057 & 0.762 $\pm$ 0.087 \\
Drug     & {\bf 0.835 $\pm$ 0.027} & 1.011 $\pm$ 0.027 & 1.072 $\pm$ 0.019 & {\bf 0.845 $\pm$ 0.033} & 0.875 $\pm$ 0.037 & 0.911 $\pm$ 0.036 & 0.844 $\pm$ 0.033 \\ \bottomrule
\end{tabular}
}
\end{table*}

\subsubsection*{Accuracy}
First, we demonstrate the predictive performances of GPX and the comparing methods in Table~\ref{tab:experiment:accuracy}.
GPX achieved the lowest predictive errors on all the datasets, compared to the other globally or locally linear models.
In addition, their predictive errors were comparable to that of GPR on all the datasets.
This result indicates that GPR can be replaced by GPX to achieve similar predictive performances.

\begin{table*}[t]
\centering
\caption{
Average and standard deviation of faithfulness scores on each dataset (higher scores are better).
This table can be interpreted similarly as Table~\ref{tab:experiment:accuracy}.
For Wine dataset, we could not measure the scores of the methods except for GPX owing to computational time limitations.
For Paper and Drug datasets, we did not evaluate the scores as changes in $\mat{Z}$ cannot reflect $\mat{X}$.
}
\label{tab:experiment:faithfulness}
\resizebox{\textwidth}{!}{%
\begin{tabular}{@{}rrrrrrrr@{}}
\toprule
& GPX (ours)              & GPR+LIME           & GPR+SHAP         & KL  & Localized         & Network     & ARD      \\ \midrule
Digits   & {\bf 0.888 $\pm$ 0.003} & 0.384 $\pm$ 0.038 & 0.651 $\pm$ 0.013 & 0.757 $\pm$ 0.022 & 0.300 $\pm$ 0.029 & 0.352 $\pm$ 0.021 & -0.059 $\pm$ 0.012 \\
Abalone  & {\bf 0.898 $\pm$ 0.017} & 0.775 $\pm$ 0.024 & {\bf 0.914 $\pm$ 0.007} & 0.551 $\pm$ 0.024 & 0.432 $\pm$ 0.042 & 0.497 $\pm$ 0.033 & 0.288 $\pm$ 0.107  \\
Diabetes & {\bf 0.966 $\pm$ 0.008} & 0.844 $\pm$ 0.027 & 0.928 $\pm$ 0.010 & 0.558 $\pm$ 0.050 & 0.340 $\pm$ 0.074 & 0.365 $\pm$ 0.086 & 0.593 $\pm$ 0.027  \\
Boston   & {\bf 0.898 $\pm$ 0.026} & 0.693 $\pm$ 0.035 & {\bf 0.869 $\pm$ 0.009} & 0.518 $\pm$ 0.080 & 0.562 $\pm$ 0.075 & 0.525 $\pm$ 0.039 & 0.195 $\pm$ 0.042  \\
Fish     & {\bf 0.902 $\pm$ 0.016} & 0.672 $\pm$ 0.043 & 0.826 $\pm$ 0.033 & 0.428 $\pm$ 0.012 & 0.480 $\pm$ 0.046 & 0.394 $\pm$ 0.090 & 0.190 $\pm$ 0.067  \\
Wine     & {\bf 0.749 $\pm$ 0.027} & 0.575 $\pm$ 0.009 & {\bf 0.734 $\pm$ 0.008} & 0.234 $\pm$ 0.028 & 0.270 $\pm$ 0.013 & 0.321 $\pm$ 0.017 & -0.006 $\pm$ 0.019 \\ \bottomrule
\end{tabular}
}
\end{table*}

\subsubsection*{Faithfulness}
Assessing the correctness of the estimated contribution of each feature to a prediction requires a reference ``true'' contribution for comparison. 
As this is rarely available, a typical approach for measuring the faithfulness of the contributions produced by interpretable models is to rely on the proxy notion of the contributions: observing the effect of removing features on the model’s prediction. 
Following previous studies~\citep{melis2018,bhatt2020evaluating}, we computed the faithfulness score by removing features one-by-one, measuring the differences between the original predictions and the predictions from the inputs without the removed features, and calculating the correlation between the differences and the contributions of the removed features.

Table~\ref{tab:experiment:faithfulness} shows the faithfulness scores of GPX and the comparing methods.
Here, we denote the results of LIME and Kernel SHAP using GPR as the black-box prediction model by GPR+LIME and GPR+SHAP, respectively.
We found that GPX achieved the best faithfulness scores on all the datasets.
As GPX predicts and explains using a single locally linear model for each test sample, when removing a feature from the input, the contribution of the feature is subtracted from the prediction directly.
Meanwhile, because GPR+LIME, GPR+SHAP and KL have different prediction and explanation models, a gap may exist between the estimated contribution in the explanation model and the latent contribution in the prediction.
Because the predictions by GPX and GPR were performed using similar calculations, their faithfulness differences were likely due to the gap.
With ARD, it cannot estimate feature relevances appropriate for each sample, as the feature relevances are shared over samples; therefore, it produced relatively low faithfulness scores.

\begin{figure*}[t]
\begin{minipage}{0.32\hsize}
\begin{center}
\includegraphics[width=\textwidth]{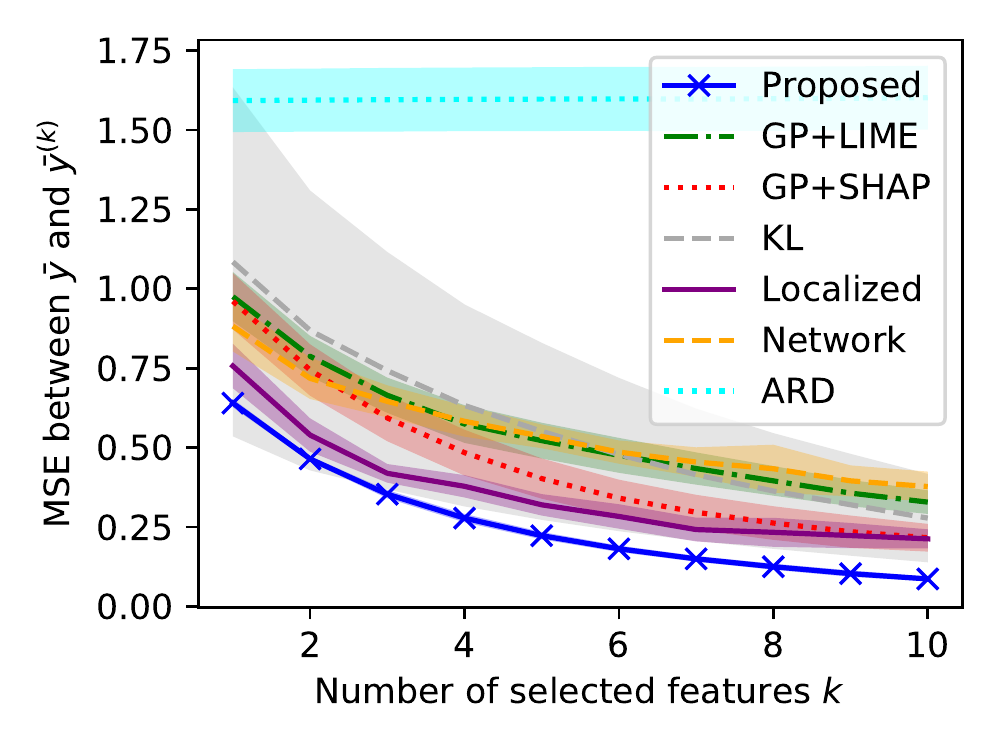}\\(a) Digits
\\
\includegraphics[width=\textwidth]{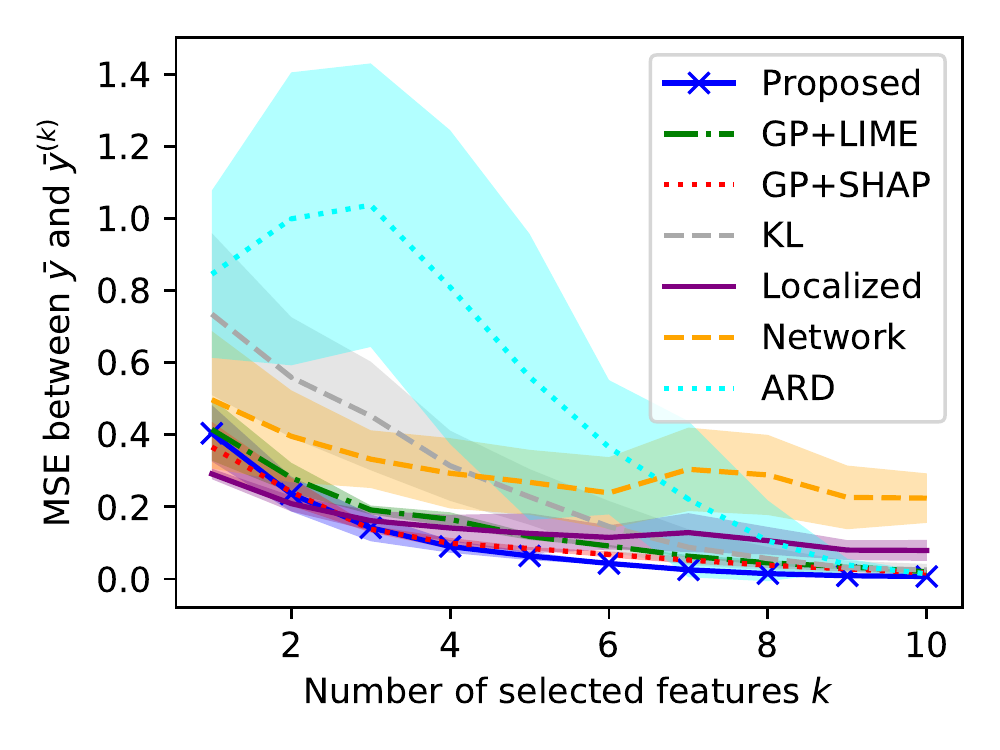}\\(d) Boston
\end{center}
\end{minipage}
\begin{minipage}{0.32\hsize}
\begin{center}
\includegraphics[width=\textwidth]{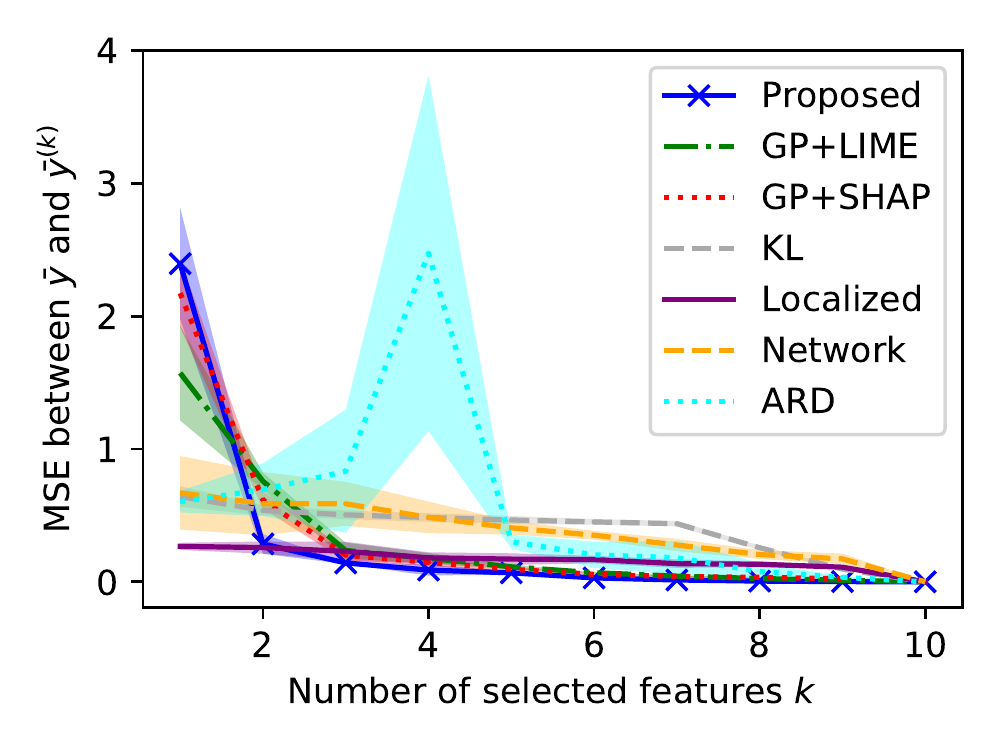}\\(b) Abalone
\\
\includegraphics[width=\textwidth]{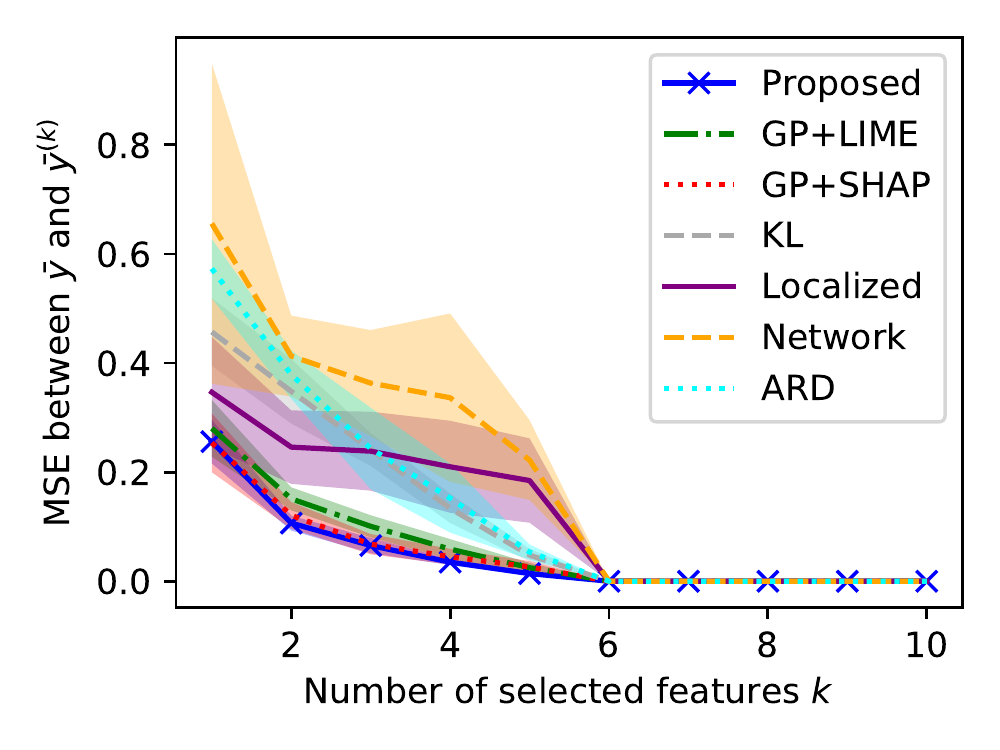}\\(e) Fish
\end{center}
\end{minipage}
\begin{minipage}{0.32\hsize}
\begin{center}
\includegraphics[width=\textwidth]{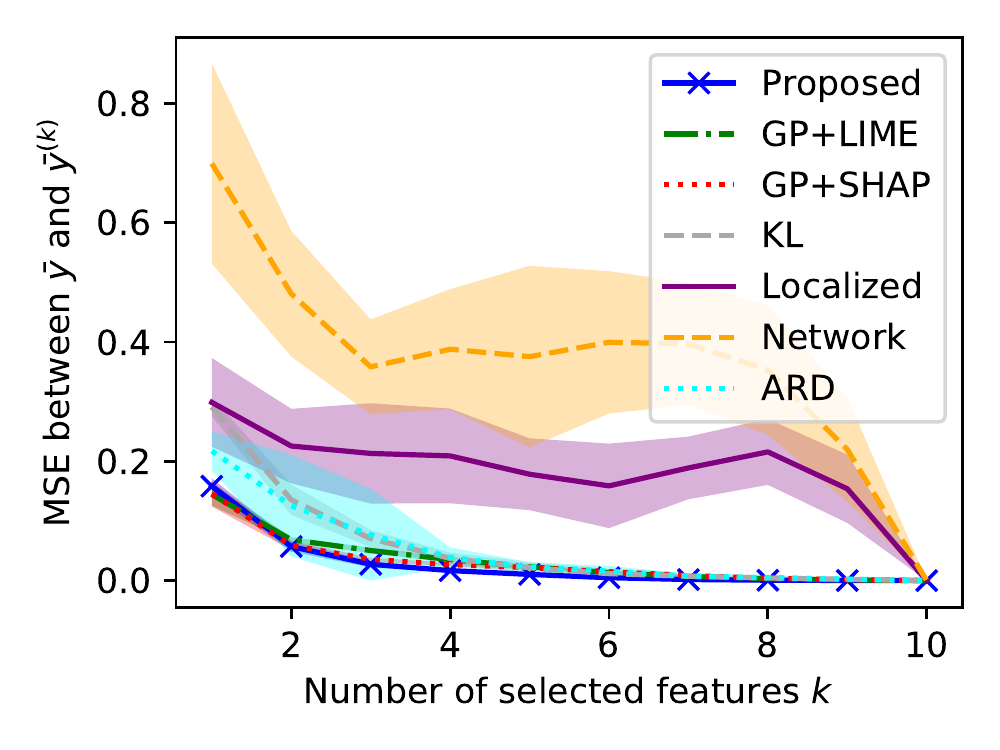}\\(c) Diabetes
\\
\includegraphics[width=\textwidth]{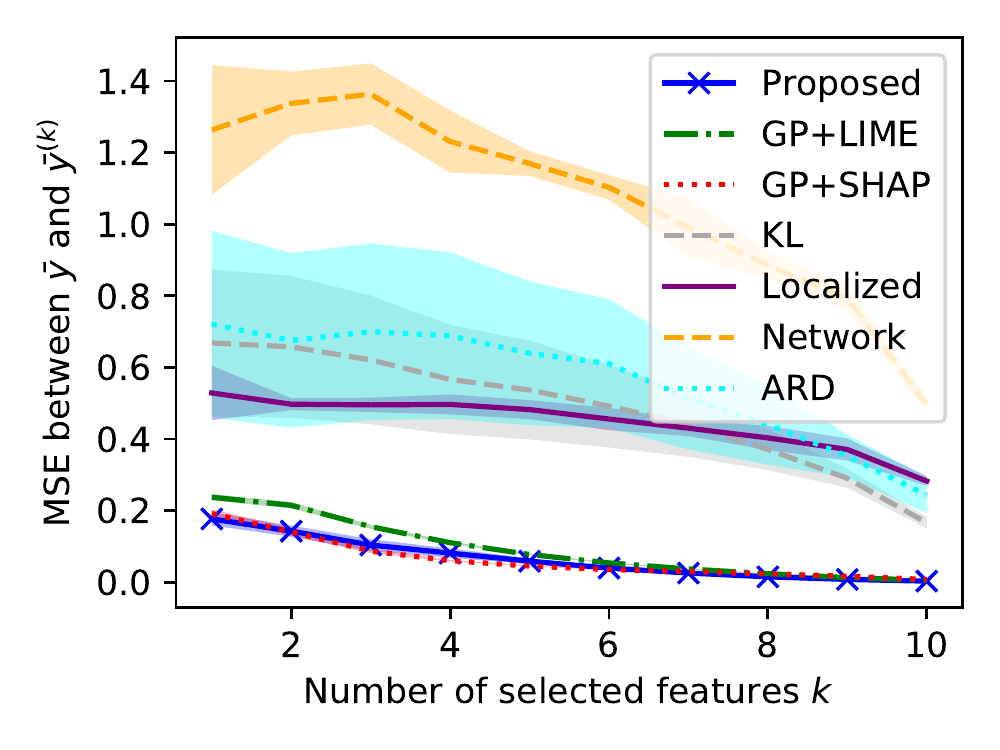}\\(f) Wine
\end{center}
\end{minipage}
\caption{Average sufficiency scores on Digits, Abalone, Diabetes, Boston, Fish and Wine datasets (lower scores are better). The filled area on each line indicates its standard deviation.
For Paper and Drug datasets, we did not evaluate the scores as changes in $\mat{Z}$ cannot reflect $\mat{X}$.
}
\label{fig:experiment:sufficiency}
\end{figure*}

\subsubsection*{Sufficiency}
In general, the inputs contain many irrelevant features that do not contribute to the predictions, and discovering important features in all the features is difficult for users of the models.
Therefore, a desirable property of the interpretable models is that it can assign high contributions only for important features that affect the predictions well.
To quantify how each method satisfies the property, we define the sufficiency score at $k$, where $k$ is the number of important features.
In particular, the sufficiency score at $k$ was computed by identifying $k$ important features in the descending order of the absolute values of their estimated contributions, predicting from the inputs having only $k$ important features, and comparing them against the original predictions.
Because the number of important features varied according to the sample and dataset, we evaluated them at $k=1,2,\cdots,10$.

Figure~\ref{fig:experiment:sufficiency} shows the sufficiency scores of GPX and the comparing methods.
Independent of the $k$ values, GPX outperformed the others on Digits dataset, whereas GPX, GPR+LIME, and GPR+SHAP produced the best sufficiency scores on Diabetes, Fish and Wine datasets.
These results indicate that GPX was appropriately assigned high contributions for the important features. 
On Abalone and Boston datasets, GPX was slightly inferior to the localized lasso at $k=1,2$, although GPX outperformed it at $k \geq 3$.
This is because the localized lasso has a regularizer that induces sparse weights.
This result suggests that GPX can be further improved by employing the mechanism for generating sparse weights.

\begin{table*}[t]
\centering
\caption{
Average and standard deviation of stability scores on each dataset (lower scores are better).
This table can be interpreted similarly as Table~\ref{tab:experiment:accuracy}.
We did not measure the scores for models that have global feature relevances/weights, such as ARD, as their scores are obviously zero according to~\bref{eq:stability}.
}
\label{tab:experiment:stability}
\resizebox{\textwidth}{!}{%
\begin{tabular}{@{}rrrrrrrr@{}}
\toprule
              & GPX (ours)        & GPR+LIME          & GPR+SHAP          & KL                  & Localized         & Network           \\ \midrule

Digits   & {\bf 1.153 $\pm$ 0.015} & 2.410 $\pm$ 0.079            & 2.274 $\pm$ 0.110  & 2.058 $\pm$ 1.043 & 1.996 $\pm$ 0.098  & 1.989 $\pm$ 0.222  \\
Abalone  & {\bf 3.094 $\pm$ 0.249} & 11.809 $\pm$ 0.453           & 11.625 $\pm$ 0.362 & 4.485 $\pm$ 0.911 & 15.439 $\pm$ 1.193 & 15.772 $\pm$ 0.948 \\
Diabetes & {\bf 1.164 $\pm$ 0.065} & 1.870 $\pm$ 0.119            & 1.400 $\pm$ 0.147  & 1.401 $\pm$ 0.113 & 2.833 $\pm$ 0.085  & 2.693 $\pm$ 0.123  \\
Boston   & {\bf 1.452 $\pm$ 0.081} & 4.180 $\pm$ 0.426            & 2.176 $\pm$ 0.304  & 2.956 $\pm$ 0.221 & 3.794 $\pm$ 1.215  & 3.467 $\pm$ 0.799  \\
Fish     & {\bf 1.497 $\pm$ 0.075} & $>10^5$  & 1.634 $\pm$ 0.057  & 3.289 $\pm$ 0.536 & 5.746 $\pm$ 0.563  & 5.625 $\pm$ 0.591  \\
Wine     & {\bf 1.531 $\pm$ 0.113} & $>10^6$  & $>10^5$            & 5.255 $\pm$ 0.906 & 3.989 $\pm$ 0.306  & 4.033 $\pm$ 0.282  \\
Paper    & {\bf 0.004 $\pm$ 0.002} & 5.535 $\pm$ 0.161            & 4.548 $\pm$ 0.098           & {\bf 0.026 $\pm$ 0.052} & 6.482 $\pm$ 0.161  & 6.986 $\pm$ 0.442  \\
Drug     & {\bf 0.067 $\pm$ 0.014} & 16.976 $\pm$ 0.399           & 12.037 $\pm$ 0.513          & 11.547 $\pm$ 0.265 & 16.908 $\pm$ 0.532 & 17.341 $\pm$ 0.589 \\ 
\bottomrule
\end{tabular}
}
\end{table*}

\subsubsection*{Stability}
To generate meaningful explanations, interpretable methods must be robust against local perturbations from the input, as explanations that are sensitive to slight changes in the input may be regarded as inconsistent by users.
In particular, flexible models such as locally linear models might be sensitive to such changes for achieving better predictions.
As with the work by~\citep{melis2018}, we used the following quantity for measuring the stability of the estimated weights for test sample $(\mat{x}_*, \mat{z}_*)$, as follows:
%
\begin{align}
\label{eq:stability}
L(\mat{x}_*, \mat{z}_*) = \max_{\mat{x}'_*, \mat{z}'_* \in \mathcal{B}_\epsilon(\mat{x}_*)} \frac{\| \mat{w}_*' - \mat{w}_* \|_2}{\| \mat{z}_*' - \mat{z}_* \|_2}, 
\quad
\mathcal{B}_\epsilon(\mat{x}_*) = \{(\mat{x}_*',\mat{z}_*') \in \mathcal{D}_\mathrm{te} \mid \frac{1}{m}\| \mat{x}_*' - \mat{x}_* \|_2 < \epsilon \}, 
\end{align}
where, $\mathcal{D}_\mathrm{te} = \{ (\mat{x}_*, \mat{z}_*) \}$ is a set of test samples; $\mat{w}_*$ and $\mat{w}_*'$ are the standardized estimated weights associated with test samples $ (\mat{x}_*, \mat{z}_*)$ and $(\mat{x}_*', \mat{z}_*')$, respectively; $\epsilon > 0$ is a parameter that determines neighboring samples; $m$ is the dimensionality of $\mat{x}_*$.
We set $\epsilon = 0.05$ in our experiments.
Intuitively, the stability score will be high when the estimated weights for the sample and its neighboring samples are similar.
Subsequently, we computed the stability score on a dataset by averaging the quantity~\bref{eq:stability} on all the test samples in the dataset.

Table~\ref{tab:experiment:stability} shows the stability scores on each dataset.
GPX achieved the best stability scores on all the datasets.
With GPR+LIME and GPR+SHAP, their stability scores were lower than that of GPX, although the prediction powers of GPX and GPR were comparable.
This would be because LIME and Kernel SHAP estimated the weights independently over the test samples.
The stability score of KL was as good as that of GPX only on Paper dataset; however, on the other datasets, KL was inferior to than GPX.

\begin{figure*}[t]
\begin{center}
\includegraphics[width=\textwidth]{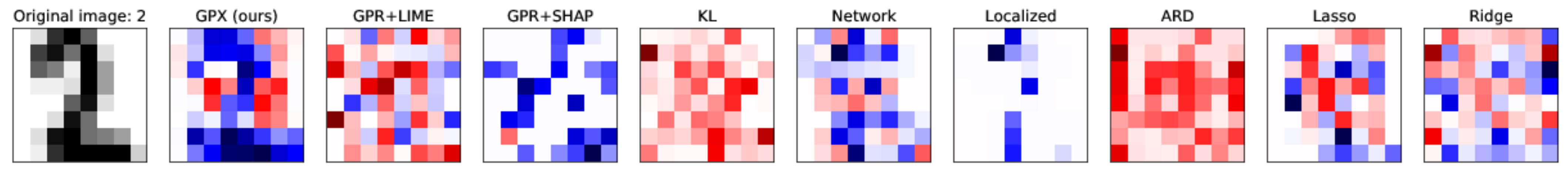}\\
\includegraphics[width=\textwidth]{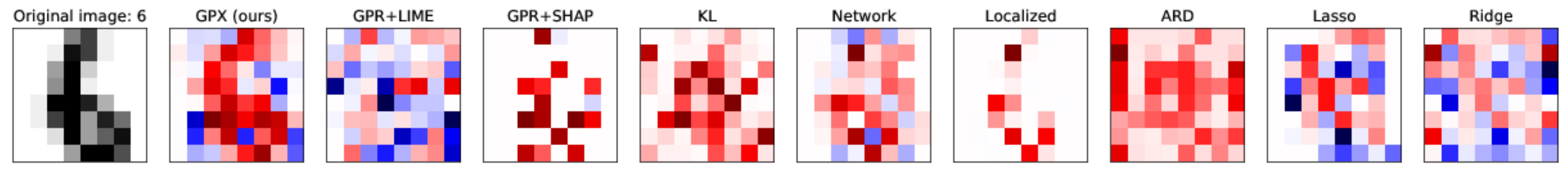}\\
\end{center}
\caption{
Examples of estimated weights of each model on Digits dataset.
The upper row shows the weights for the sample with digit two ($y=-1$), whereas the bottom one displays those for the sample with digit six ($y=1$).
Red and blue denote positive and negative weights, respectively, and their color strengths represent their magnitudes.
}
\label{fig:experiment:digits}
\end{figure*}

\subsubsection*{Qualitative comparison}
Finally, we qualitatively compared the estimated weights using GPX and the comparing methods on Digits dataset, in which the appropriate contributions for predictions were apparent.
For this comparison, we rescaled the inputs $\mat{X}$ and $\mat{Z}$ to be within $[0, 1]$.

Figure~\ref{fig:experiment:digits} shows the estimated weights on two samples.
We provide the results for all the digits in Appendix~\ref{sec:appendix:digits}.
On this dataset, the appropriate weights can be obtained by assigning weights having the same sign with the target variable to black pixels.
We found that the methods except for GPX and the localized lasso could not estimate reasonable weights.
Meanwhile, the weights estimated by GPX and the localized lasso were appropriate, although they exhibited different characteristics, i.e., dense weights from GPX, whereas sparse ones from the localized lasso.
The task determines the better explanation; however, as showing important regions rather than pixels is meaningful for images, the estimated weights using GPX would be easier to interpret on Digits dataset.
Furthermore, the degree of sparsity in the localized lasso can be changed as a hyperparameter; if the value of the hyperparameter is zero, the localized lasso is identical to the network lasso.
However, because the estimated weights using the network lasso were inappropriate, those using GPX cannot be mimicked by the localized lasso.

\begin{table*}[t]
\centering
\caption{
Training, prediction and total times (seconds) of each method on the Digits dataset.
The training times of GPR+LIME/SHAP and KL are those of GPR and ARD, respectively.
The prediction times of GPR+LIME/SHAP and KL are those of producing explanations for all the test samples.
}
\label{tab:experiment:time}
\begin{tabular}{@{}rrrrrrrr@{}}
\toprule
              & GPX (ours)   & GPR+LIME & GPR+SHAP & KL       & Localized & Network & ARD  \\ \midrule
Training   & 5.74  & 5.53     & 5.76     & 6.57     & 105.84    & 106.16  & 6.18 \\
Prediction & 24.57 & 155.54   & 2643.06  & 156.19   & 1.59      & 1.61    & 0.27 \\
Total      & 30.31 & 161.07   & 2648.82  & 162.76   & 107.43    & 107.77  & 6.45 \\ \bottomrule
\end{tabular}
\end{table*}

\subsubsection*{Computational efficiency}
Table~\ref{tab:experiment:time} shows the computational times of each of the methods on Digits dataset.
First, the training time of GPX was much the same as those of GPR and ARD, and significantly faster than those of the localized and network lasso.
Since the localized and network lasso requires the hyperparameter search, the their actual training times were about 48 and 12 times longer than the times shown in the table, respectively.
Second, the prediction time of GPX was significantly faster than those of GPR+LIME/SHAP and KL.
This is because GPX does not require learning model parameters at prediction phase.
On the other hand, since GPR+LIME/SHAP learns an explanation model for each sample at that time, and KL requires producing a lot of predictions with slight changes in the value of each dimension of the input, their prediction times lead to increase.

\section{Conclusion}\label{sec:conclusion}
We proposed a GP-based regression model with sample-wise explanations.
The proposed model assumes that each sample has a locally linear model, which is used for both prediction and explanation, and the weight vector of the locally linear model are generated from multivariate GP priors.
The hyperparameters of the proposed models were estimated by maximizing the marginal likelihood, in which all the weight vectors were integrated out.
Subsequently, for a test sample, the proposed model predicted its target variable and weight vector with uncertainty.
In the experiments, we confirmed that the proposed model outperformed the existing globally and locally linear models and achieved comparable performances with the standard GPR in terms of predictive performance, and the proposed model was superior to the existing methods, including model-agnostic interpretable methods, in terms of three interpretability measurements.
Then, we showed that the feature weights estimated by the proposed model were appropriate as the explanation.

In future studies, we will confirm the effectiveness of the proposed model by applying its concept into various problems in which GPs have been successfully used, such as classification, black-box optimization, and time-series forecasting.
In addition, we will extend the proposed model for further improvements in interpretability, e.g., by employing the mechanism of inducing sparsity for the weight vectors.

\section*{Acknowledgment}
This work was supported by JSPS KAKENHI Grant Number 18K18112.

\appendix

\section{Feature Description and Additional Examples for the Boston Housing Dataset}
\label{sec:appendix:boston}

\begin{table}[t]
\centering
\caption{Feature names and their descriptions for the Boston housing dataset.}
\label{tab:appendix:boston}
\begin{tabular}{@{}rp{65mm}@{}}
\toprule
Feature name & Description                                                         \\ \midrule
CRIM         & Per capita crime rate by town                                       \\
ZN           & Proportion of residential land zoned for lots over 25,000 sq.ft.    \\
INDUS        & Proportion of non-retail business acres per town.                   \\
CHAS         & Charles River dummy variable (1 if tract bounds river; 0 otherwise) \\
NOX          & Nitric oxides concentration (parts per 10 million)                  \\
RM           & Average number of rooms per dwelling                                \\
AGE          & Proportion of owner-occupied units built prior to 1940              \\
DIS          & Weighted distances to five Boston employment centers                \\
RAD          & Index of accessibility to radial highways                           \\
TAX          & Full-value property-tax rate per \$10,000                            \\
PTRATIO      & Pupil-teacher ratio by town                                        \\
B            & 1000(Bk - 0.63)\textasciicircum 2 where Bk is the proportion of blacks by town      \\
LSTAT        & \% lower status of the population                                    \\
MEDV         & Median value of owner-occupied homes in \$1000's                     \\ \bottomrule
\end{tabular}
\end{table}

The Boston housing dataset, referred to as ``Boston'' in our experiments, contains information collected by the U.S. Census Service regarding housing in the area of Boston, Massachusetts~\citep{harrison1978hedonic} and is used for predicting house prices based on the information.
Table~\ref{tab:appendix:boston} lists the names of the features and their descriptions for the Boston housing dataset.

\begin{figure*}[t]
\begin{minipage}{0.49\hsize}
\begin{center}
\includegraphics[width=0.7\textwidth]{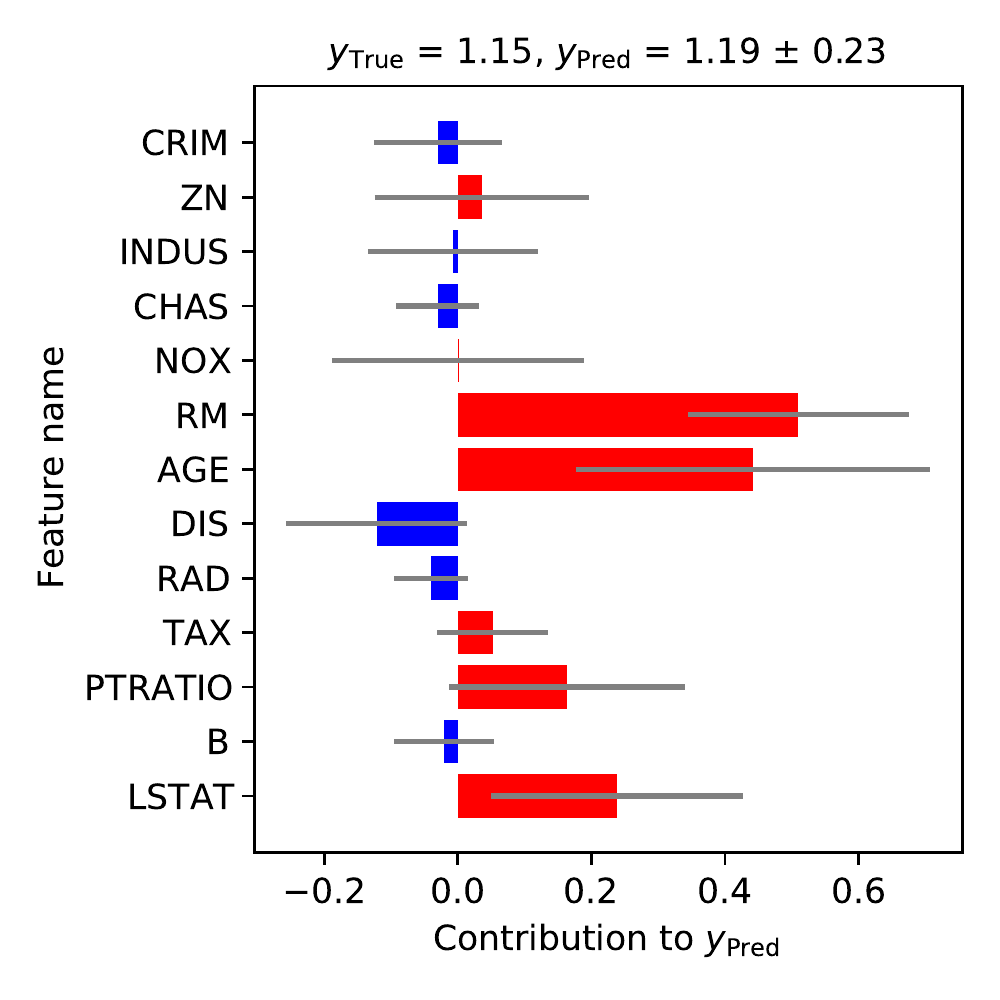}\\
\includegraphics[width=0.7\textwidth]{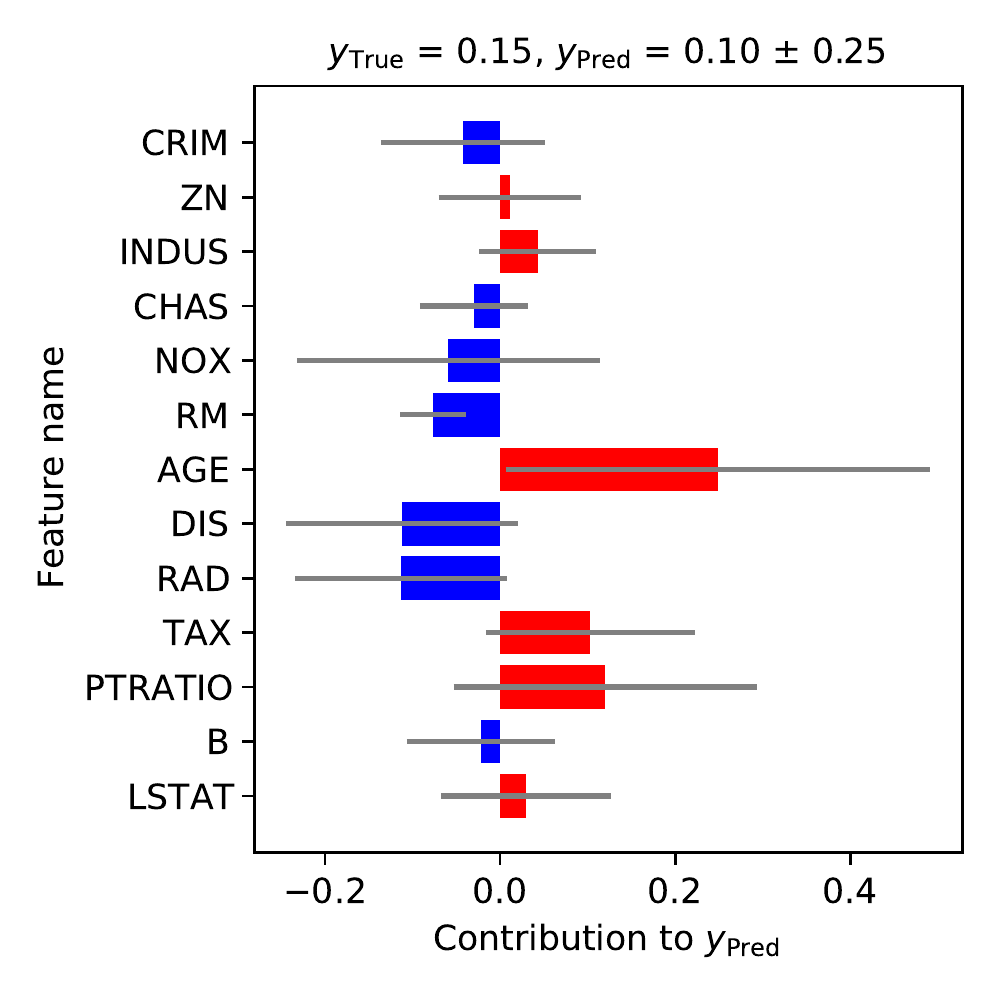}
\end{center}
\end{minipage}
\begin{minipage}{0.49\hsize}
\begin{center}
\includegraphics[width=0.7\textwidth]{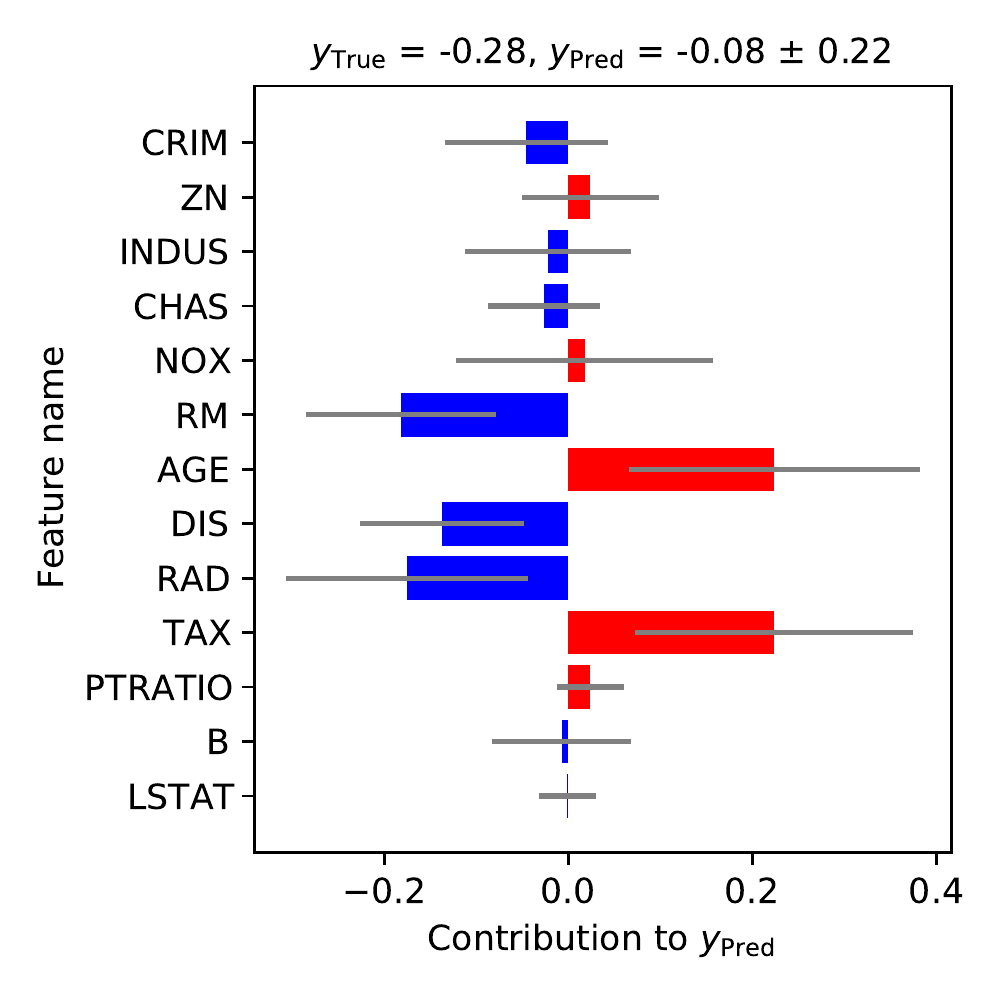}\\
\includegraphics[width=0.7\textwidth]{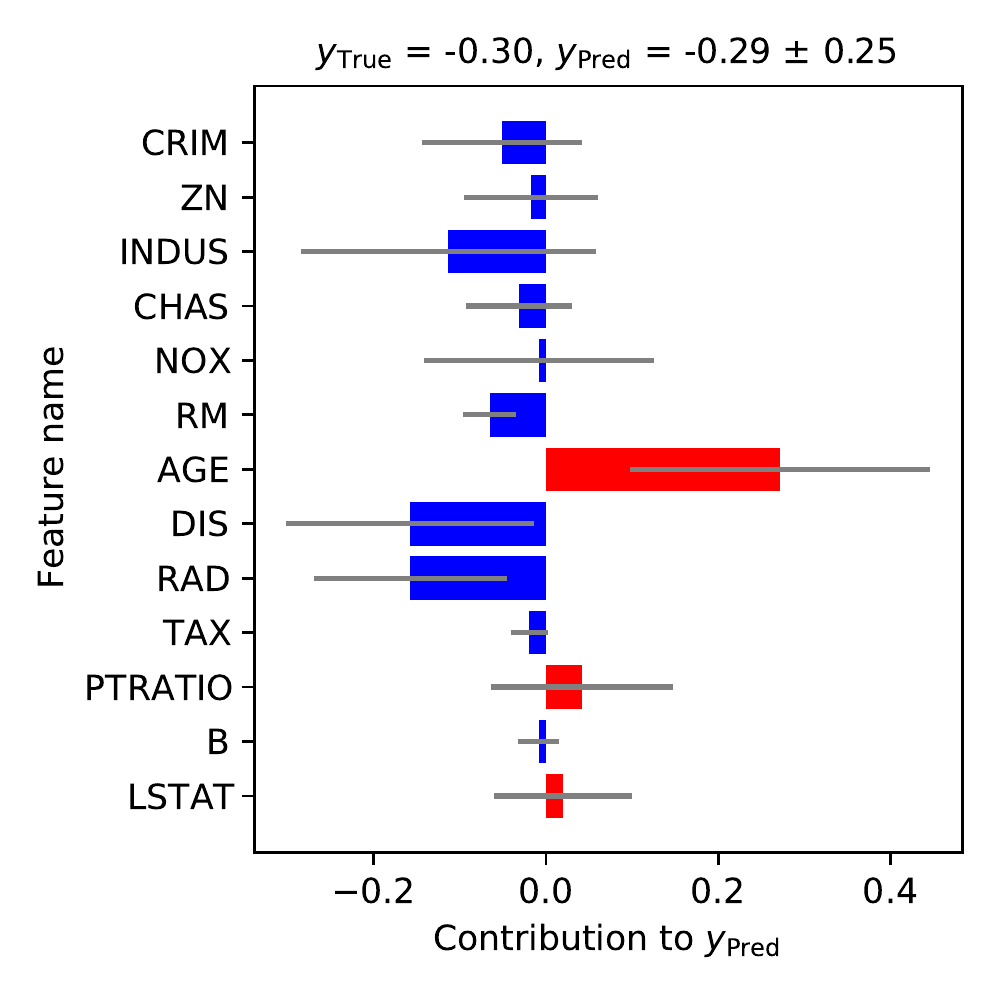}
\end{center}
\end{minipage}
\caption{
Examples of feature contributions estimated by GPX for the Boston housing dataset.
Here, the error bars denote the standard deviations or feature contributions.
}
\label{fig:appendix:boston_results}
\end{figure*}

Figure~\ref{fig:appendix:boston_results} presents four examples of feature contributions estimated by GPX.
We found that each of these examples has different feature contributions, although some of the features, such as ``AGE'' and ``DIS,'' had consistent positive or negative contributions, respectively.

\section{Detailed Derivation of Predictive Distributions}
\label{sec:appendix:derivation}

In this appendix, we describe the derivation of predictive distributions in detail.
For a new test sample $(\mat{x}_*, \mat{z}_*)$, our goal is to infer the predictive distributions of the target variable $y_*$ and weight vector $\mat{w}_*$.

\subsubsection*{Predictive distribution of $y_*$}
The predictive distribution of $y_*$ is obtained similarly to the standard GPR~\citep{rasmussen2003gaussian}.
In Section 4, we demonstrated that the marginal distribution of training target variables $\mat{y}$ for GPX is defined as
\begin{align}
\label{eq:appendix:marginal}
p(\mat{y} \mid \mat{X}, \mat{Z}) 
&= \iint p(\mat{y}, \mat{W}, \mat{G} \mid \mat{X}, \mat{Z}) d\mat{W} d\mat{G} \\
&= \mathcal{N}\left(\mat{y} \mid \mat{0},\mat{C} \right), \nonumber
\end{align}
where $\mat{C} = \sigma_\mathrm{y}^2\mat{I}_n + (\mat{K}  + \sigma_\mathrm{w}^2 \mat{I}_n) \odot \mat{Z}\mat{Z}^\top$.
According to~\bref{eq:appendix:marginal}, the joint marginal distribution of $\mat{y}$ and $y_*$ is defined as
\begin{equation}
p(\mat{y}, y_* \mid \mat{X},\mat{Z},\mat{x}_*,\mat{z}_*)
= \mathcal{N}\left(
\left[ \begin{array}{c} \mat{y} \\ y_*\end{array} \right] 
\Bigg\vert\ \mathbf{0},\left[\begin{array}{cc}
    \mat{C} & \mat{c}_* \\
    \mat{c}_*^\top & c_{**}
    \end{array}\right]\right),
\label{eq:appendix:joint_marginal}
\end{equation}
where $\mat{c}_* = \left(k_\theta(\mat{x}_*, \mat{x}_i) \mat{z}_*^\top \mat{z}_i \right)_{i=1}^n \in \mathbb{R}^n$, and $c_{**} = \sigma_\mathrm{y}^2 + \left(k_\theta(\mat{x}_*, \mat{x}_*) + \sigma_\mathrm{w}^2 \right)\mat{z}_*^\top \mat{z}_*  \in \mathbb{R}$.
The predictive distribution of $y_*$ is the conditional distribution of $y_*$ given $\mat{y}$ with training and testing inputs. Therefore, it can be obtained by applying the formula of conditional distributions for normal distributions~\cite[Eq.~(354)]{Petersen2008} to~\bref{eq:appendix:joint_marginal} as follows:
\begin{equation}
\label{eq:appendix:predictive_y}
p(y_* \mid \mat{x}_*,\mat{z}_*,\set{D})
= \mathcal{N}\left(y_* \mid \mat{c}_*^\top\mat{C}^{-1}\mat{y}, c_{**} - \mat{c}_*^\top \mat{C}^{-1} \mat{c}_* \right).
\end{equation}

\subsubsection*{Predictive distribution of $\mat{w}_*$}
The predictive distribution of $\mat{w}_*$ can be obtained by solving the following equation:
\begin{equation}
p(\mat{w}_* \mid \mat{x}_*,\mat{z}_*,\set{D}) 
= \int p(\mat{w}_* \mid \mat{W},\mat{X},\mat{x}_*) p(\mat{W} \mid \set{D}) d\mat{W},
\label{eq:appendix:predictive_w1}
\end{equation}
where the first integrand $p(\mat{w}_* \mid \mat{W},\mat{X},\mat{x}_*)$ is the conditional distribution of $\mat{w}_*$
and the second integrand $p(\mat{W} \mid \set{D})$ is the posterior distribution of $\mat{W}$.
The conditional distribution of $\mat{w}_*$ is derived similarly to the conditional distribution of $y_*$~\bref{eq:appendix:predictive_y}.
The distribution of $\mat{W}$ in which the functions $\mat{G}$ are integrated out is given by
%
\begin{align}
p(\mat{W} \mid \mat{X}) &= \int p(\mat{W} \mid \mat{G}) p(\mat{G} \mid \mat{X}) d\mat{G} \\
&= \prod_{l=1}^d \mathcal{N}\left(\mat{W}_{\cdot,l} \mid \mat{0}, \mat{K} + \sigma_\mathrm{w}^2 \mat{I}_n \right), \nonumber
\end{align}
where $\mat{W}_{\cdot,l}$ is the $l$th column vector of $\mat{W}$.
According to this, the joint distribution of $\mat{W}$ and $\mat{w}_*$ is defined as
%
\begin{align}
\label{eq:appendix:joint_marginal_w}
p(\mat{W},\mat{w}_* \mid \mat{X},\mat{x}_*)
= \prod_{l=1}^d \mathcal{N}\left( 
\left[\begin{array}{c}\mat{W}_{\cdot,l} \\ w_{*,l} \end{array}\right]\ \Big\vert\ 
\mat{0}, \left[ \begin{array}{cc} \mat{K} + \sigma_\mathrm{w}^2 \mat{I}_n & \mat{k}_* \\ \mat{k}_*^\top & k_{**} \end{array} \right] 
\right), 
\end{align}
where we let $\mat{k}_* = \left(k_\theta(\mat{x}_*,\mat{x}_i) \right)_{i=1}^n$ and $k_{**} = k_\theta(\mat{x}_*,\mat{x}_*) + \sigma_\mathrm{w}^2$.
Subsequently, we can obtain the conditional distribution of $\mat{w}_*$ by applying the formula of conditional distributions for normal distributions~\cite[Eq.~(354)]{Petersen2008} to~\bref{eq:appendix:joint_marginal_w} as follows:
%
\begin{align}
\label{eq:appendix:conditional_w1}
\lefteqn{p(\mat{w}_* \mid \mat{W},\mat{X},\mat{x}_*)}\quad\\
&= \prod_{l=1}^d \mathcal{N}\Big(
w_{*,l}\ \Big\vert\ \mat{k}_*^\top \left(\mat{K} + \sigma_\mathrm{w}^2\mat{I}_n \right)^{-1} \mat{W}_{\cdot,l}, 
k_{**} - \mat{k}_*^\top \left(\mat{K} + \sigma_\mathrm{w}^2\mat{I}_n \right)^{-1} \mat{k}_* 
\Big). \nonumber
\end{align}
Here, we can rewrite \bref{eq:appendix:conditional_w1} as a single $d$-dimensional multivariate normal distribution as follows:
%
\begin{align}
\lefteqn{p(\mat{w}_* \mid \mat{W},\mat{X},\mat{x}_*)}\quad\\
&= \mathcal{N}\Big(\mat{w}_*\ \Big\vert\ \barmat{k}_*^\top\left(\barmat{K} + \sigma_\mathrm{w}^2 \mat{I}_{nd} \right)^{-1}\rmvec(\mat{W}),
\barmat{c}_{**} - \barmat{k}_*^\top\left(\barmat{K} + \sigma_\mathrm{w}^2 \mat{I}_{nd} \right)^{-1}\barmat{k}_* \Big), \nonumber
\end{align}
where $\bar{\mat{K}}$ is a block diagonal matrix of order $nd$ whose block is $\mat{K}$, $\rmvec(\cdot)$ is a function that flattens the input matrix in column-major order, and $\barmat{c}_{**} = \left(k_\theta(\mat{x}_*,\mat{x}_*) + \sigma_\mathrm{w}^2 \right)\mat{I}_d$.
$\barmat{k}_*$ is an $nd$-by-$d$ block matrix, where each block is an $n$-by-$1$ matrix the and $(l,l)$-block of the block matrix is $(k_\theta(\mat{x}_*,\mat{x}_i))_{i=1}^n$ for $l=1,2,\cdots,d$, while the other blocks are zero matrices.
By letting $\mat{A} = \barmat{k}_*^\top\left(\barmat{K} + \sigma_\mathrm{w}^2 \mat{I}_{nd} \right)^{-1}$, we obtain 
\begin{equation}
p(\mat{w}_* \mid \mat{W},\mat{X},\mat{x}_*)
= \mathcal{N}(\mat{w}_* \mid \mat{A}\rmvec(\mat{W}), \barmat{c}_{**} - \mat{A}\barmat{k}_* ).
\label{eq:appendix:conditional_w2}
\end{equation}
To derive the posterior distribution of $\mat{W}$, $p(\mat{W} \mid \set{D})$, we first consider the joint distribution of $\mat{W}$ and $\set{D}$.
This distribution is straightforwardly obtained as 
\begin{equation}
p(\mat{W},\set{D})
= \prod_{l=1}^d \mathcal{N}\left(\mat{W}_{\cdot,l} \mid \mat{0}, \mat{K} \right) \prod_{i=1}^n \mathcal{N}(y_i \mid \mat{w}_i^\top \mat{z}_i, \sigma_\mathrm{y}^2),
\end{equation}
which can be rewritten as 
\begin{equation}
p(\mat{W},\set{D})
= \mathcal{N}\left(\rmvec(\mat{W}) \mid \mat{0}, \barmat{K} \right) 
\mathcal{N}\left(\mat{y} \mid \barmat{Z}\rmvec(\mat{W}), \sigma_\mathrm{y}^2 \mat{I}_n \right),
\label{eq:appendix:posterior_w1}
\end{equation}
where $\barmat{Z} = (\diag(\mat{Z}_{\cdot,1}), \diag(\mat{Z}_{\cdot,2}), \cdots, \diag(\mat{Z}_{\cdot,d})) \in \mathbb{R}^{n \times nd}$.
By applying the formula of conditional distributions of normal distributions~\cite[Eqs. (2.113)--(2.117)]{bishop2006pattern} to~\bref{eq:appendix:posterior_w1}, we can obtain 
\begin{align}
p(\mat{W} \mid \set{D}) = \mathcal{N}(\rmvec(\mat{W}) \mid \sigma_\mathrm{y}^{-2} \mat{\Sigma}\barmat{Z}^\top \mat{y}, \mat{\Sigma}),\\
\intertext{where}
\mat{\Sigma} = \left(\barmat{K}^{-1} - \barmat{Z}^\top (\sigma_\mathrm{y}^{-2} \mat{I}_n) \barmat{Z} \right)^{-1}.
\end{align}
Here, the computation of $\mat{\Sigma}$ requires inverting a square matrix of order $nd$ with a computational complexity of $\set{O}(n^3d^3)$.
By using the Woodbury identity~\cite[Eq. (156)]{Petersen2008} to compute this inversion efficiently, we can transform $\mat{\Sigma}$ into $\mat{S} = \barmat{K} - \barmat{K}\barmat{Z}^\top \mat{D}^{-1} \barmat{Z}\barmat{K}$, which requires inverting a matrix of order $n$, $\mat{D} = \sigma_\mathrm{y}^2 \mat{I}_n + \mat{K} \odot \mat{Z}\mat{Z}^\top$.
Consequently, we obtain 
\begin{equation}
p(\mat{W} \mid \set{D}) = \mathcal{N}(\rmvec(\mat{W}) \mid \sigma_\mathrm{y}^{-2} \mat{S}\barmat{Z}^\top \mat{y}, \mat{S}).
\label{eq:appendix:posterior_w2}
\end{equation}

From~\bref{eq:appendix:conditional_w2} and~\bref{eq:appendix:posterior_w2}, one can see that \bref{eq:appendix:predictive_w1} can be represented by the following equation:
%
\begin{align}
p(\mat{w}_* \mid \mat{x}_*,\mat{z}_*,\set{D})
= \int \mathcal{N}(\mat{w}_* \mid \mat{A}\rmvec(\mat{W}), \barmat{c}_{**} - \mat{A}\barmat{k}_* ) 
\mathcal{N}(\rmvec(\mat{W}) \mid \sigma_\mathrm{y}^{-2} \mat{S}\barmat{Z}^\top \mat{y}, \mat{S}) d\mat{W}.
\end{align}
This integral can be obtained in a closed form, as shown in~\cite[Eqs. (2.113)--(2.117)]{bishop2006pattern}.
Therefore, we can obtain the predictive distribution of $\mat{w}_*$ as follows:
%
\begin{align}
\label{eq:appendix:predictive_w}
p(\mat{w}_* \mid \mat{x}_*,\mat{z}_*,\set{D})
= \mathcal{N}(\mat{w}_* \mid \sigma_\mathrm{y}^{-2} \mat{A} \mat{S} \barmat{Z}^\top \mat{y}, \barmat{c}_{**} - \mat{A}\barmat{k}_* + \mat{A}\mat{S}\mat{A}^\top). 
\end{align}

\section{Specification of Datasets}
\label{sec:appendix:dataset}

We considered eight datasets from the UCI machine learning repository~\citep{dua2019}, which were referred to as Digits\footnote{\url{https://archive.ics.uci.edu/ml/datasets/Optical+Recognition+of+Handwritten+Digits}}, Abalone\footnote{\url{https://archive.ics.uci.edu/ml/datasets/Abalone}}, Diabetes\footnote{\url{https://archive.ics.uci.edu/ml/datasets/diabetes}}, Boston~\citep{harrison1978hedonic}, Fish\footnote{\url{https://archive.ics.uci.edu/ml/datasets/QSAR+fish+toxicity}}, Wine\footnote{\url{https://archive.ics.uci.edu/ml/datasets/wine+quality}}, Paper\footnote{\url{https://archive.ics.uci.edu/ml/datasets/Paper+Reviews}}, and Drug\footnote{\url{https://archive.ics.uci.edu/ml/datasets/Drug+Review+Dataset+(Druglib.com)}} in our experiments.

\begin{table}[t]
\centering
\caption{Specification of datasets.}
\label{tab:appendix:dataset}
\begin{tabular}{@{}rrrrrrrrr@{}}
\toprule
    & Digits & Abalone & Diabetes & Boston & Fish & Wine  & Paper & Drug  \\ \midrule
$n$ & 1,797  & 4,177   & 442      & 506    & 908  & 6,497 & 399   & 3,989 \\
$d$ & 64     & 10      & 10       & 13     & 6    & 11    & 2,990 & 2,429 \\ \bottomrule
\end{tabular}
\end{table}

The first six datasets consisted of tabular data. We treated the original inputs $\mat{X}$ and simplified inputs $\mat{Z}$ identically in our experiments.
The Digits dataset was originally developed as a classification dataset for recognizing handwritten digits from zero to nine.
As described in Section~5.1, we used this dataset for a regression problem by transforming the digit labels into binary values of 1 or $-1$.
Here, we used only the testing set from the original Digits dataset because that is how scikit-learn~\citep{pedregosa2011} distributes this dataset.
The Abalone dataset is a dataset for predicting the age of abalone based on physical measurements.
The Diabetes dataset is a dataset for predicting the onset of diabetes based on diagnostic measures.
The Boston dataset is a dataset for predicting house prices, as described in Appendix~\ref{sec:appendix:boston}.
The Fish dataset is a dataset for predicting acute aquatic toxicity toward the fish pimephales promelas for a set of chemicals.
The Wine dataset is a dataset for predicting the quality of white and red wines based on physicochemical tests.
The remaining two datasets are text datasets.
The Paper dataset is a dataset for predicting evaluation scores for papers based on review texts written mainly in Spanish.
The Drug dataset is a drug review dataset for predicting 10-star ratings for drugs based on patient review texts.
For each dataset, $\mat{X}$ and $\mat{Z}$ are different. Specifically, we used the 512-dimensional sentence vectors obtained using sentence transformers~\citep{reimers2020making} as $\mat{X}$ and used bag-of-words binary vectors of the sentences as $\mat{Z}$, where the cutoff frequencies for words were set to two and five for the Paper and Drug datasets, respectively.
Table~\ref{tab:appendix:dataset} lists the number of samples $n$ and number of features $d$ in each dataset.

\section{Detailed Description of Comparing Methods}
\label{sec:appendix:method}
In this appendix, we describe the implementation and hyperparameter search methods used for comparing methods.

We implemented GPR using PyTorch v1.5.0\footnote{\url{https://pytorch.org/}}.
All hyperparameters for GPR were estimated by maximizing marginal likelihood~\citep{rasmussen2003gaussian}, where we initialized the hyperparameters to the same values as those for GPX.
For Lasso and Ridge, we used the implementations provided by scikit-learn~\citep{pedregosa2011}.
The hyperparameters that regularize the strengths of the $\ell_1$ and $\ell_2$ regularizers in Lasso and Ridge, respectively, were optimized through a grid search using functions provided by scikit-learn (i.e., \texttt{sklearn.linear\_model.LassoCV} and \texttt{sklearn.linear\_model.RidgeCV}) with the default options. The search range for the hyperparameters for Lasso was limited to within 100 grid points such that the ratio of its minimum value to its maximum value was capped at 0.001, while that for Ridge was limited to within a range of $\{0.1, 1, 10 \}$.
For ARD, we implemented it as with GPR.
Herein, the kernel function used in ARD is defined as
\begin{equation}
k_{\mat{\theta}}(\mat{x}, \mat{x}') = \alpha \exp\left( - \frac{1}{2} \sum_{l=1}^d (x_l - x'_l)^2 / \theta_l \right)\quad\ (\alpha, \theta_l > 0),
\end{equation}
where $\alpha$ is a scale parameter; $\theta_l$ is the relevance of the $l$th feature; $\mat{\theta} = \{ \alpha, \theta_1, \theta_2, \cdots, \theta_d \}$ is a set of parameters to be estimated as with GPR.
For the localized lasso, we used the original implementation written in Python\footnote{\url{https://riken-yamada.github.io/localizedlasso.html}}.
The hyperparameters and their search ranges for the localized lasso are the strength of network regularization $\lambda_1 \in \{1, 3, 5, 7\}$, strength of the $\ell_{1,2}$ regularizer $\lambda_2 \in \{0.01, 0.1, 1, 10\}$, and $k \in \{5, 10, 15\}$ for the $k$-nearest-neighbor graph.
The hyperparameters were optimized through a grid search.
The network lasso is a special case of the localized lasso. If $\lambda_2$ for the localized lasso is zero, then the localized lasso is identical to the network lasso.
Therefore, we used the implementation of the localized lasso and set $\lambda_2 = 0$ for the network lasso.
The hyperparameter search for the network lasso was the same as that for the localized lasso, except for the setting of $\lambda_2$.
For LIME and Kernel SHAP, we used the original implementations\footnote{LIME: \url{https://github.com/marcotcr/lime}, Kernel SHAP: \url{https://github.com/slundberg/shap}}.
For KL, we implemented it by mimicking its original implementation\footnote{\url{https://github.com/topipa/gp-varsel-kl-var}}.
Herein, we set to amount of perturbation $\Delta=0.001$ throughout our experiments.

\begin{figure*}[h!]
\begin{center}
\includegraphics[width=\textwidth]{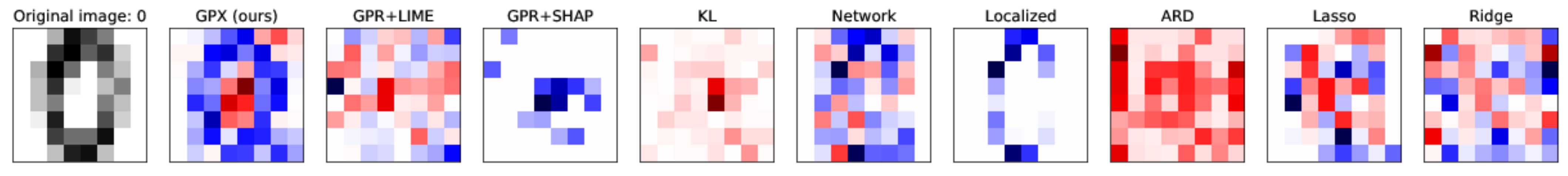}\\
\includegraphics[width=\textwidth]{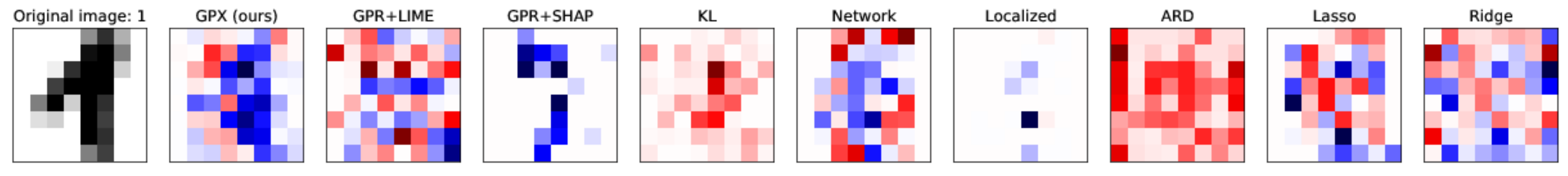}\\
\includegraphics[width=\textwidth]{fig/visualize_digits-idx_2-all-label_2.pdf}\\
\includegraphics[width=\textwidth]{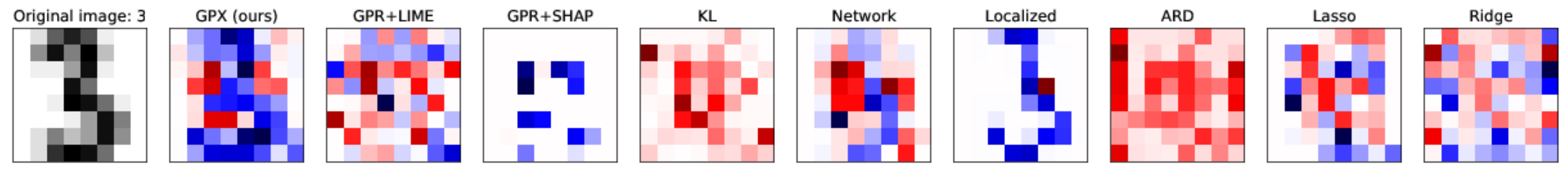}\\
\includegraphics[width=\textwidth]{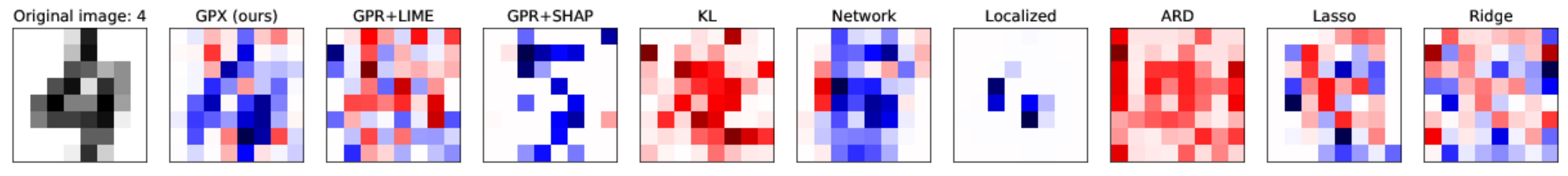}\\
\includegraphics[width=\textwidth]{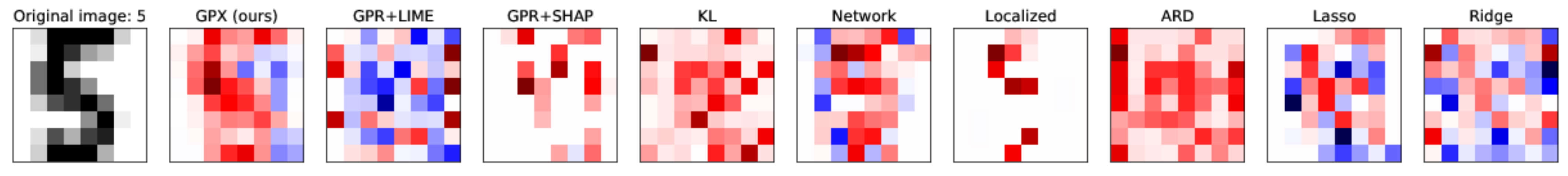}\\
\includegraphics[width=\textwidth]{fig/visualize_digits-idx_14-all-label_6.pdf}\\
\includegraphics[width=\textwidth]{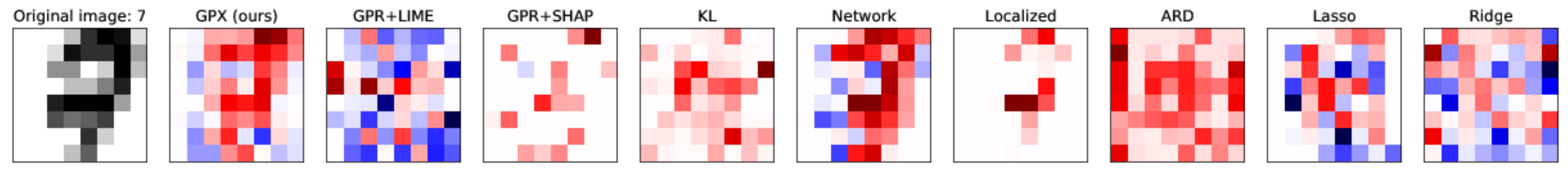}\\
\includegraphics[width=\textwidth]{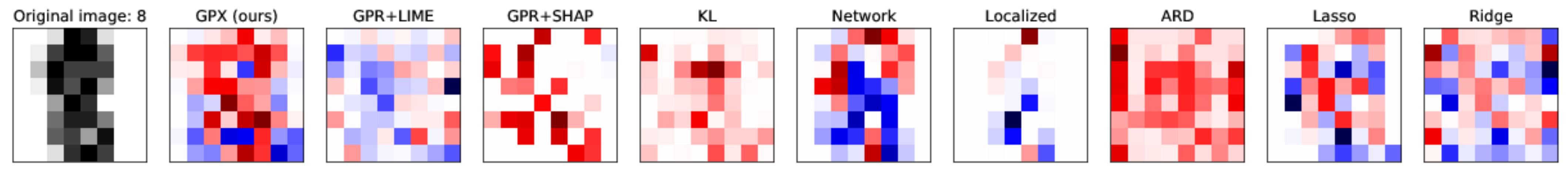}\\
\includegraphics[width=\textwidth]{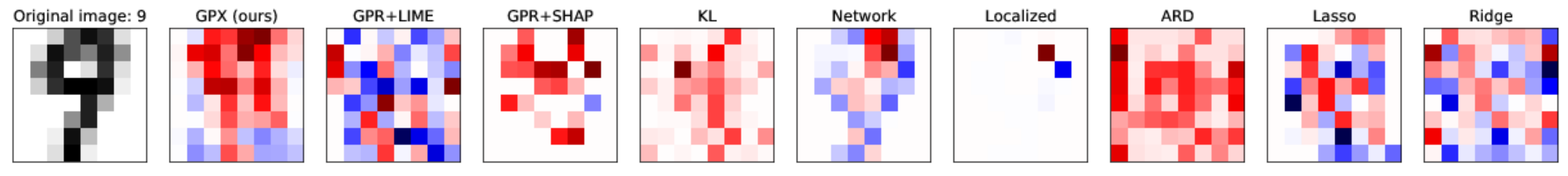}
\end{center}
\caption{
Examples of estimated weights for digits ranging from zero to nine for the Digits dataset.
The five upper rows present the weights of samples with digits of zero to four ($y=-1$), whereas the five bottom rows present those for samples with digits from five to nine ($y=1$).
Red and blue denote positive and negative weights, respectively, and their color strengths represent their magnitudes.
}
\label{fig:appendix:digits}
\end{figure*}

\section{Additional Results for the Digits Dataset}
\label{sec:appendix:digits}


Figure~\ref{fig:appendix:digits} presents additional examples of estimated weights for the Digits dataset.
We found that the weights estimated by GPX were appropriately assigned such that the regions of black pixels have weights with the same signs as those of the target variables.

\begin{figure*}[h!]
\begin{center}
\includegraphics[width=\textwidth]{fig/visualize_digits-idx_2-all-label_2.pdf}\\
\includegraphics[width=\textwidth]{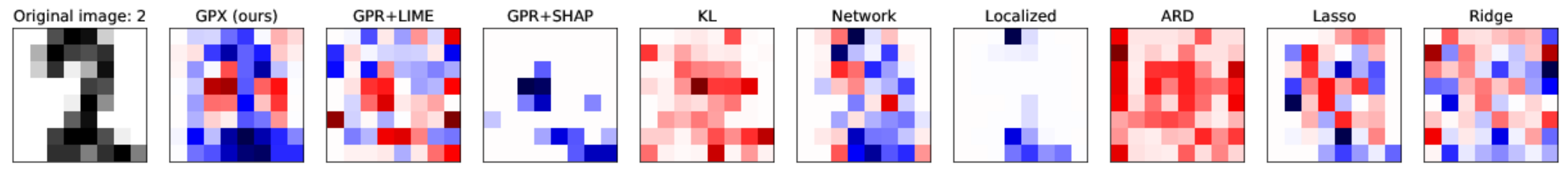}\\
\includegraphics[width=\textwidth]{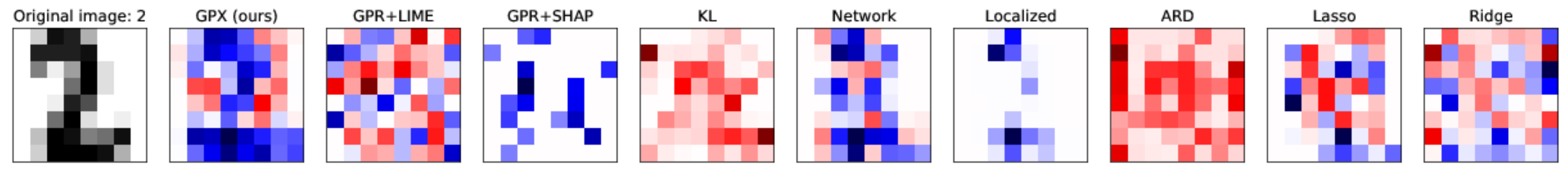}
\end{center}
\caption{
Different examples of estimated weights for digit two for the Digits dataset.
}
\label{fig:appendix:digits_stability}
\end{figure*}

In terms of the stability of explanations, estimated weights for the same digit should be similar.
Figure~\ref{fig:appendix:digits_stability} presents three examples of estimated weights for the digit two.
We found that GPX estimated similar weights for all three examples.


\bibliographystyle{unsrtnat}
\bibliography{gpx}  

\begin{thebibliography}{43}
\providecommand{\natexlab}[1]{#1}
\providecommand{\url}[1]{\texttt{#1}}
\expandafter\ifx\csname urlstyle\endcsname\relax
  \providecommand{\doi}[1]{doi: #1}\else
  \providecommand{\doi}{doi: \begingroup \urlstyle{rm}\Url}\fi

\bibitem[Rasmussen(2003)]{rasmussen2003gaussian}
Carl~Edward Rasmussen.
\newblock Gaussian processes in machine learning.
\newblock In \emph{Summer School on Machine Learning}, pages 63--71. Springer,
  2003.

\bibitem[Wilson et~al.(2012)Wilson, Knowles, and
  Ghahramani]{wilson2012gaussian}
Andrew~Gordon Wilson, David~A Knowles, and Zoubin Ghahramani.
\newblock Gaussian process regression networks.
\newblock In \emph{Proceedings of the 29th International Coference on
  International Conference on Machine Learning}, pages 1139--1146, 2012.

\bibitem[Csat{\'o} et~al.(2000)Csat{\'o}, Fokou{\'e}, Opper, Schottky, and
  Winther]{csato2000efficient}
Lehel Csat{\'o}, Ernest Fokou{\'e}, Manfred Opper, Bernhard Schottky, and Ole
  Winther.
\newblock Efficient approaches to gaussian process classification.
\newblock In \emph{Advances in neural information processing systems}, pages
  251--257, 2000.

\bibitem[Roberts et~al.(2013)Roberts, Osborne, Ebden, Reece, Gibson, and
  Aigrain]{roberts2013gaussian}
Stephen Roberts, Michael Osborne, Mark Ebden, Steven Reece, Neale Gibson, and
  Suzanne Aigrain.
\newblock Gaussian processes for time-series modelling.
\newblock \emph{Philosophical Transactions of the Royal Society A:
  Mathematical, Physical and Engineering Sciences}, 371\penalty0
  (1984):\penalty0 20110550, 2013.

\bibitem[Snoek et~al.(2012)Snoek, Larochelle, and Adams]{snoek2012practical}
Jasper Snoek, Hugo Larochelle, and Ryan~P Adams.
\newblock Practical bayesian optimization of machine learning algorithms.
\newblock In \emph{Advances in neural information processing systems}, pages
  2951--2959, 2012.

\bibitem[Gonzalvez et~al.(2019)Gonzalvez, Lezmi, Roncalli, and
  Xu]{gonzalvez2019financial}
Joan Gonzalvez, Edmond Lezmi, Thierry Roncalli, and Jiali Xu.
\newblock Financial applications of gaussian processes and bayesian
  optimization.
\newblock \emph{arXiv preprint arXiv:1903.04841}, 2019.

\bibitem[Camps-Valls et~al.(2016)Camps-Valls, Verrelst, Munoz-Mari, Laparra,
  Mateo-Jimenez, and Gomez-Dans]{camps2016survey}
Gustau Camps-Valls, Jochem Verrelst, Jordi Munoz-Mari, Valero Laparra, Fernando
  Mateo-Jimenez, and Jose Gomez-Dans.
\newblock A survey on gaussian processes for earth-observation data analysis: A
  comprehensive investigation.
\newblock \emph{IEEE Geoscience and Remote Sensing Magazine}, 4\penalty0
  (2):\penalty0 58--78, 2016.

\bibitem[Zhang et~al.(2020)Zhang, Apley, and Chen]{zhang2020bayesian}
Yichi Zhang, Daniel~W Apley, and Wei Chen.
\newblock Bayesian optimization for materials design with mixed quantitative
  and qualitative variables.
\newblock \emph{Scientific Reports}, 10\penalty0 (1):\penalty0 1--13, 2020.

\bibitem[Cheng et~al.(2017)Cheng, Darnell, Dumitrascu, Chivers, Draugelis, Li,
  and Engelhardt]{cheng2017sparse}
Li-Fang Cheng, Gregory Darnell, Bianca Dumitrascu, Corey Chivers, Michael~E
  Draugelis, Kai Li, and Barbara~E Engelhardt.
\newblock Sparse multi-output gaussian processes for medical time series
  prediction.
\newblock \emph{arXiv preprint arXiv:1703.09112}, 2017.

\bibitem[Futoma(2018)]{futoma2018gaussian}
Joseph Futoma.
\newblock \emph{Gaussian process-based models for clinical time series in
  healthcare}.
\newblock PhD thesis, Duke University, 2018.

\bibitem[Molnar(2019)]{molnar2019}
Christoph Molnar.
\newblock \emph{Interpretable Machine Learning}.
\newblock 2019.
\newblock \url{https://christophm.github.io/interpretable-ml-book/}.

\bibitem[Chen et~al.(2018)Chen, Song, Wainwright, and Jordan]{chen2018learning}
Jianbo Chen, Le~Song, Martin Wainwright, and Michael Jordan.
\newblock Learning to explain: An information-theoretic perspective on model
  interpretation.
\newblock In \emph{International Conference on Machine Learning}, pages
  883--892, 2018.

\bibitem[Ribeiro et~al.(2016)Ribeiro, Singh, and Guestrin]{ribeiro2016}
Marco~Tulio Ribeiro, Sameer Singh, and Carlos Guestrin.
\newblock "{{Why Should I Trust You}}?": {{Explaining}} the {{Predictions}} of
  {{Any Classifier}}.
\newblock In \emph{Proceedings of the {{22Nd ACM SIGKDD International
  Conference}} on {{Knowledge Discovery}} and {{Data Mining}}}, {{KDD}} '16,
  pages 1135--1144. {ACM}, 2016.
\newblock ISBN 978-1-4503-4232-2.
\newblock \doi{10.1145/2939672.2939778}.

\bibitem[Lundberg and Lee(2017)]{lundberg2017}
Scott~M Lundberg and Su-In Lee.
\newblock A {{Unified Approach}} to {{Interpreting Model Predictions}}.
\newblock In I.~Guyon, U.~V. Luxburg, S.~Bengio, H.~Wallach, R.~Fergus,
  S.~Vishwanathan, and R.~Garnett, editors, \emph{Advances in {{Neural
  Information Processing Systems}} 30}, pages 4765--4774. {Curran Associates,
  Inc.}, 2017.

\bibitem[Harrison~Jr and Rubinfeld(1978)]{harrison1978hedonic}
David Harrison~Jr and Daniel~L Rubinfeld.
\newblock Hedonic housing prices and the demand for clean air.
\newblock \emph{Journal of Environmental Economics and Management}, 5:\penalty0
  81--102, 1978.

\bibitem[{\'A}lvarez et~al.(2012){\'A}lvarez, Rosasco, and
  Lawrence]{alvarez2012kernels}
Mauricio~A {\'A}lvarez, Lorenzo Rosasco, and Neil~D Lawrence.
\newblock Kernels for vector-valued functions: A review.
\newblock \emph{Foundations and Trends{\textregistered} in Machine Learning},
  4\penalty0 (3):\penalty0 195--266, 2012.

\bibitem[Hoerl and Kennard(1970)]{hoerl1970}
Arthur~E Hoerl and Robert~W Kennard.
\newblock Ridge regression: {{Biased}} estimation for nonorthogonal problems.
\newblock \emph{Technometrics}, 12\penalty0 (1):\penalty0 55--67, 1970.

\bibitem[Tibshirani(1996)]{tibshirani1996}
Robert Tibshirani.
\newblock Regression shrinkage and selection via the lasso.
\newblock \emph{Journal of the Royal Statistical Society: Series B
  (Methodological)}, 58\penalty0 (1):\penalty0 267--288, 1996.

\bibitem[Neal(2012)]{neal2012bayesian}
Radford~M Neal.
\newblock \emph{Bayesian learning for neural networks}, volume 118.
\newblock Springer Science \& Business Media, 2012.

\bibitem[Wipf and Nagarajan(2008)]{wipf2008new}
David~P Wipf and Srikantan~S Nagarajan.
\newblock A new view of automatic relevance determination.
\newblock In \emph{Advances in neural information processing systems}, pages
  1625--1632, 2008.

\bibitem[Hallac et~al.(2015)Hallac, Leskovec, and Boyd]{hallac2015}
David Hallac, Jure Leskovec, and Stephen Boyd.
\newblock Network {{Lasso}}: {{Clustering}} and {{Optimization}} in {{Large
  Graphs}}.
\newblock In \emph{Proceedings of the 21th {{ACM SIGKDD International
  Conference}} on {{Knowledge Discovery}} and {{Data Mining}}}, {{KDD}} '15,
  pages 387--396. {ACM}, 2015.
\newblock ISBN 978-1-4503-3664-2.
\newblock \doi{10.1145/2783258.2783313}.

\bibitem[Yamada et~al.(2017)Yamada, Koh, Iwata, {Shawe-Taylor}, and
  Kaski]{yamada2017}
Makoto Yamada, Takeuchi Koh, Tomoharu Iwata, John {Shawe-Taylor}, and Samuel
  Kaski.
\newblock Localized {{Lasso}} for {{High}}-{{Dimensional Regression}}.
\newblock In Aarti Singh and Jerry Zhu, editors, \emph{Proceedings of the 20th
  {{International Conference}} on {{Artificial Intelligence}} and
  {{Statistics}}}, volume~54 of \emph{Proceedings of {{Machine Learning
  Research}}}, pages 325--333. {PMLR}, April 2017.

\bibitem[Paananen et~al.(2019)Paananen, Piironen, Andersen, and
  Vehtari]{paananen2019variable}
Topi Paananen, Juho Piironen, Michael~Riis Andersen, and Aki Vehtari.
\newblock Variable selection for gaussian processes via sensitivity analysis of
  the posterior predictive distribution.
\newblock In \emph{The 22nd International Conference on Artificial Intelligence
  and Statistics}, pages 1743--1752, 2019.

\bibitem[Chen et~al.(2019)Chen, Li, Tao, Barnett, Rudin, and Su]{chen2019looks}
Chaofan Chen, Oscar Li, Daniel Tao, Alina Barnett, Cynthia Rudin, and
  Jonathan~K Su.
\newblock This looks like that: deep learning for interpretable image
  recognition.
\newblock In \emph{Advances in Neural Information Processing Systems}, pages
  8928--8939, 2019.

\bibitem[Arras et~al.(2017)Arras, Horn, Montavon, M{\"u}ller, and
  Samek]{arras2017relevant}
Leila Arras, Franziska Horn, Gr{\'e}goire Montavon, Klaus-Robert M{\"u}ller,
  and Wojciech Samek.
\newblock " what is relevant in a text document?": An interpretable machine
  learning approach.
\newblock \emph{PloS one}, 12\penalty0 (8), 2017.

\bibitem[Ying et~al.(2019)Ying, Bourgeois, You, Zitnik, and
  Leskovec]{ying2019gnnexplainer}
Zhitao Ying, Dylan Bourgeois, Jiaxuan You, Marinka Zitnik, and Jure Leskovec.
\newblock Gnnexplainer: Generating explanations for graph neural networks.
\newblock In \emph{Advances in Neural Information Processing Systems}, pages
  9240--9251, 2019.

\bibitem[Melis and Jaakkola(2018)]{melis2018}
David~Alvarez Melis and Tommi Jaakkola.
\newblock Towards robust interpretability with self-explaining neural networks.
\newblock In \emph{Advances in Neural Information Processing Systems}, pages
  7775--7784, 2018.

\bibitem[Schwab et~al.(2019)Schwab, Miladinovic, and Karlen]{schwab2019}
Patrick Schwab, Djordje Miladinovic, and Walter Karlen.
\newblock Granger-causal attentive mixtures of experts: {{Learning}} important
  features with neural networks.
\newblock In \emph{Proceedings of the {{AAAI Conference}} on {{Artificial
  Intelligence}}}, volume~33, pages 4846--4853, 2019.

\bibitem[Yoshikawa and Iwata(2020)]{yoshikawa2020neural}
Yuya Yoshikawa and Tomoharu Iwata.
\newblock Neural generators of sparse local linear models for achieving both
  accuracy and interpretability.
\newblock \emph{arXiv preprint arXiv:2003.06441}, 2020.

\bibitem[Golovin et~al.(2017)Golovin, Solnik, Moitra, Kochanski, Karro, and
  Sculley]{golovin2017google}
Daniel Golovin, Benjamin Solnik, Subhodeep Moitra, Greg Kochanski, John Karro,
  and D~Sculley.
\newblock Google vizier: A service for black-box optimization.
\newblock In \emph{Proceedings of the 23rd ACM SIGKDD international conference
  on knowledge discovery and data mining}, pages 1487--1495, 2017.

\bibitem[Vishwanathan et~al.(2010)Vishwanathan, Schraudolph, Kondor, and
  Borgwardt]{vishwanathan2010graph}
S~Vichy~N Vishwanathan, Nicol~N Schraudolph, Risi Kondor, and Karsten~M
  Borgwardt.
\newblock Graph kernels.
\newblock \emph{Journal of Machine Learning Research}, 11\penalty0
  (Apr):\penalty0 1201--1242, 2010.

\bibitem[Muandet et~al.(2012)Muandet, Fukumizu, Dinuzzo, and
  Sch{\"o}lkopf]{muandet2012learning}
Krikamol Muandet, Kenji Fukumizu, Francesco Dinuzzo, and Bernhard
  Sch{\"o}lkopf.
\newblock Learning from distributions via support measure machines.
\newblock In \emph{Advances in neural information processing systems}, pages
  10--18, 2012.

\bibitem[Yoshikawa et~al.(2014)Yoshikawa, Iwata, and
  Sawada]{yoshikawa2014latent}
Yuya Yoshikawa, Tomoharu Iwata, and Hiroshi Sawada.
\newblock Latent support measure machines for bag-of-words data classification.
\newblock In \emph{Advances in neural information processing systems}, pages
  1961--1969, 2014.

\bibitem[Liu and Nocedal(1989)]{liu1989limited}
Dong~C Liu and Jorge Nocedal.
\newblock On the limited memory bfgs method for large scale optimization.
\newblock \emph{Mathematical programming}, 45\penalty0 (1-3):\penalty0
  503--528, 1989.

\bibitem[Titsias(2009)]{titsias2009variational}
Michalis Titsias.
\newblock Variational learning of inducing variables in sparse gaussian
  processes.
\newblock In \emph{Artificial Intelligence and Statistics}, pages 567--574,
  2009.

\bibitem[Wilson and Nickisch(2015)]{wilson2015kernel}
Andrew Wilson and Hannes Nickisch.
\newblock Kernel interpolation for scalable structured gaussian processes
  (kiss-gp).
\newblock In \emph{International Conference on Machine Learning}, pages
  1775--1784, 2015.

\bibitem[Dua and Graff(2017)]{dua2019}
Dheeru Dua and Casey Graff.
\newblock {UCI} machine learning repository, 2017.
\newblock URL \url{http://archive.ics.uci.edu/ml}.

\bibitem[Reimers and Gurevych(2020)]{reimers2020making}
Nils Reimers and Iryna Gurevych.
\newblock Making monolingual sentence embeddings multilingual using knowledge
  distillation.
\newblock \emph{arXiv preprint arXiv:2004.09813}, 2020.
\newblock \url{https://github.com/UKPLab/sentence-transformers}.

\bibitem[Garreau et~al.(2017)Garreau, Jitkrittum, and
  Kanagawa]{garreau2017large}
Damien Garreau, Wittawat Jitkrittum, and Motonobu Kanagawa.
\newblock Large sample analysis of the median heuristic.
\newblock \emph{arXiv preprint arXiv:1707.07269}, 2017.

\bibitem[Bhatt et~al.(2020)Bhatt, Weller, and Moura]{bhatt2020evaluating}
Umang Bhatt, Adrian Weller, and Jos{\'e}~MF Moura.
\newblock Evaluating and aggregating feature-based model explanations.
\newblock \emph{arXiv preprint arXiv:2005.00631}, 2020.

\bibitem[Petersen and Pedersen(2012)]{Petersen2008}
K.~B. Petersen and M.~S. Pedersen.
\newblock The matrix cookbook, 2012.
\newblock URL \url{https://www.math.uwaterloo.ca/~hwolkowi/matrixcookbook.pdf}.
\newblock Version: November 15, 2012.

\bibitem[Bishop(2006)]{bishop2006pattern}
Christopher~M Bishop.
\newblock \emph{Pattern recognition and machine learning}.
\newblock springer, 2006.

\bibitem[Pedregosa et~al.(2011)Pedregosa, Varoquaux, Gramfort, Michel, Thirion,
  Grisel, Blondel, Prettenhofer, Weiss, Dubourg, et~al.]{pedregosa2011}
Fabian Pedregosa, Ga{\"e}l Varoquaux, Alexandre Gramfort, Vincent Michel,
  Bertrand Thirion, Olivier Grisel, Mathieu Blondel, Peter Prettenhofer, Ron
  Weiss, Vincent Dubourg, et~al.
\newblock Scikit-learn: Machine learning in python.
\newblock \emph{Journal of machine learning research}, 12\penalty0
  (Oct):\penalty0 2825--2830, 2011.

\end{thebibliography}

\end{document}